\pdfoutput=1

\documentclass[11pt]{article}
 
\usepackage[]{acl}

\usepackage{times}
\usepackage{latexsym}
\usepackage{xspace}
\usepackage[T1]{fontenc}

\usepackage[utf8]{inputenc}
\usepackage{microtype}
\usepackage[export]{adjustbox}
\usepackage{inconsolata}
\usepackage{amsmath}
\usepackage{graphicx}
\usepackage{booktabs}
\usepackage{multirow}
\usepackage{amsmath}
\usepackage{enumitem}

\usepackage{graphicx}
\usepackage{subcaption}
\usepackage[breakable]{tcolorbox}

\newcommand{\ours}{\textsl{OS-Genesis}\xspace}

\definecolor{veronica-red}{RGB}{196,30,58}
\definecolor{ForestGreen}{RGB}{34,139,34}
\definecolor{BrickRed}{rgb}{.72,0,0}
\definecolor{LakeBlue}{RGB}{0,61,153}
\usepackage{algorithm}
\usepackage{algpseudocode}

\definecolor{lightblue}{RGB}{68,14,196}

\usepackage{fdsymbol}
\usepackage{fontawesome}
\newcommand{\fstar}{\textsuperscript{\fontsize{6pt}{6pt}\selectfont \faStarO}}
\newcommand{\fmoon}{\textsuperscript{\fontsize{6pt}{6pt}\selectfont \faMoonO}}

\newcommand{\atree}{\texttt{a11ytree}\xspace}
\newcommand{\gpt}{\texttt{GPT-4o}\xspace}

\title{\raisebox{0.075em}{\includegraphics[scale=.07, valign=c]{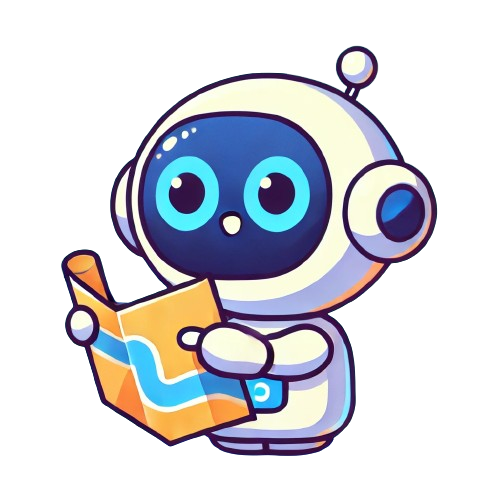}}\ours: Automating GUI Agent Trajectory Construction via Reverse Task Synthesis}

\author{
Qiushi Sun\textsuperscript{$\diamondsuit\heartsuit$}\thanks{~Equal contribution.} \quad 
Kanzhi Cheng\textsuperscript{$\diamondsuit$}\footnotemark[1] \quad 
Zichen Ding\textsuperscript{$\diamondsuit$}\footnotemark[1] \quad 
Chuanyang Jin\textsuperscript{$\clubsuit$}\footnotemark[1] \quad 
Yian Wang\textsuperscript{$\spadesuit$} \\
\textbf{Fangzhi Xu}\textsuperscript{$\diamondsuit$} \quad
\textbf{Zhenyu Wu}\textsuperscript{$\diamondsuit\spadesuit$} \quad
\textbf{Chengyou Jia}\textsuperscript{$\diamondsuit$} \quad
\textbf{Liheng Chen}\textsuperscript{$\heartsuit$} \quad
\textbf{Zhoumianze Liu}\textsuperscript{$\diamondsuit$}\\
\textbf{Ben Kao}\textsuperscript{$\heartsuit$} \quad
\textbf{Guohao Li}\fstar \quad
\textbf{Junxian He}\fmoon \quad
\textbf{Yu Qiao}\textsuperscript{$\diamondsuit$} \quad
\textbf{Zhiyong Wu}\textsuperscript{$\diamondsuit$}\\
\textsuperscript{$\diamondsuit$}Shanghai AI Laboratory 
\textsuperscript{$\heartsuit$}The University of Hong Kong \\
\textsuperscript{$\clubsuit$}Johns Hopkins University 
\textsuperscript{$\spadesuit$}Shanghai Jiao Tong University \\
\fstar University of Oxford 
\fmoon Hong Kong University of Science and Technology \\
\texttt{qiushisun@connect.hku.hk} \quad
\texttt{wuzhiyong@pjlab.org.cn} \\
}

\begin{document}

\maketitle

\begin{abstract}
Graphical User Interface (GUI) agents powered by Vision-Language Models (VLMs) have demonstrated 
human-like computer control capability.
Despite their utility in advancing digital automation, a critical bottleneck persists: collecting high-quality trajectory data for training.
Common practices for collecting such data rely on human supervision or synthetic data generation through executing pre-defined tasks, 
which are either resource-intensive or unable to guarantee data quality.
Moreover, these methods suffer from limited data diversity and significant gaps between synthetic data and real-world environments.
To address these challenges, we propose \ours, a novel GUI data synthesis pipeline that reverses the conventional trajectory collection process.
Instead of relying on pre-defined tasks, \ours enables agents first to perceive environments and perform step-wise interactions,
then retrospectively derive high-quality tasks to enable trajectory-level exploration.
A trajectory reward model is then employed to ensure the quality 
of the generated trajectories.  
We demonstrate that training GUI agents with \ours significantly improves their performance on highly challenging online benchmarks.
In-depth analysis further validates \ours's efficiency and its superior data quality and diversity compared to existing synthesis methods.
Our codes, data, and checkpoints are available at \href{https://qiushisun.github.io/OS-Genesis-Home/}{OS-Genesis Homepage}.

\end{abstract}

\section{Introduction}

\begin{figure}[ht]
\large
\centering
\includegraphics[scale=0.5]{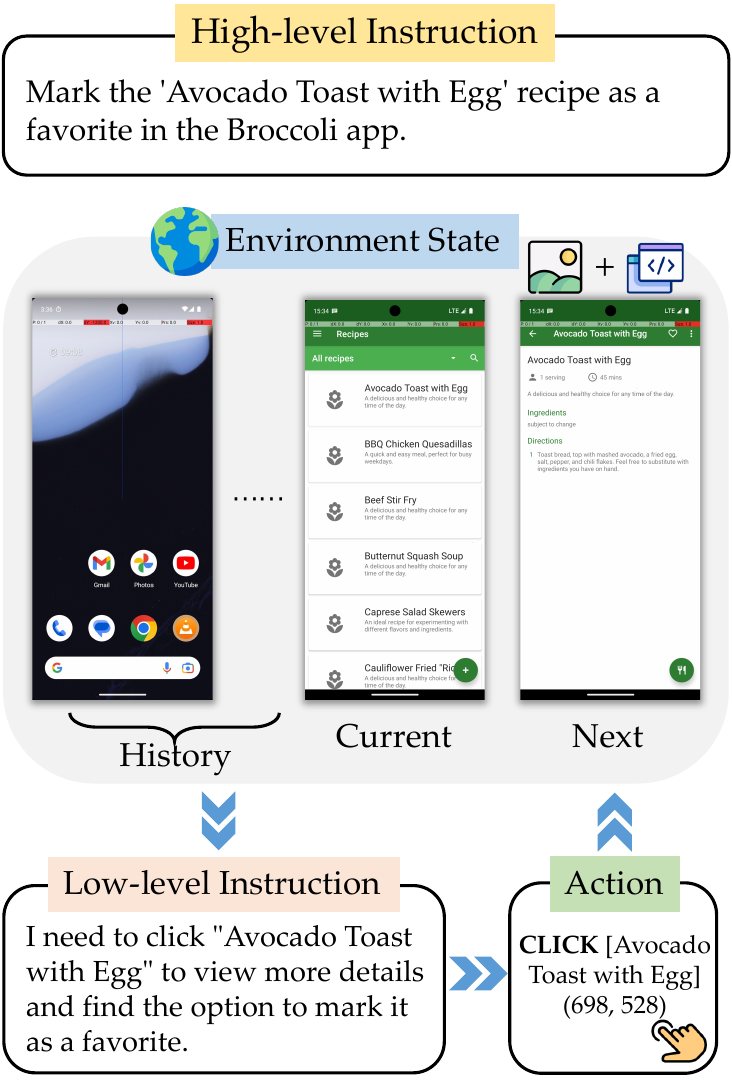}
\caption{Ideal GUI trajectory format, including High-Level Instructions, States (visual + textual representation), Low-Level Instructions, and Actions.}
\label{fig:trajectory}
\vspace{-1.25em}
\end{figure}

Recent advancements in Vision-Language Models~(VLMs; \citealp{chen2023internvl,Qwen2VL}) have driven researchers to build a variety of language agents~\citep{sumers2024cognitive}. 
As an emerging class of AI systems, 
these agents are being explored for their potential to automate complicated computer tasks on Graphical User Interfaces~(GUIs), 
aiming to achieve digital automation~\citep{claude2023,hu2024dawnguiagent}.
To complete GUI tasks autonomously,
an agent must possess key capabilities:
understanding user intentions, planning tasks, and executing actions.
Therefore, 
using high-quality trajectories for training is essential for improving their agentic capabilities~\citep{zheng2024synapse}.

As illustrated in Figure~\ref{fig:trajectory},
ideal GUI agent trajectories contain the following key components:
(1) a high-level instruction that defines the overall goal the agent aims to accomplish, 
(2) a series of low-level instructions that each describe specific steps required,
(3) actions (\textit{e.g.}, \texttt{CLICK}, \texttt{TYPE})
and 
(4) states, which include visual representations like screenshots and
textual representations such as \atree\footnote{\atree: Accessibility (a11y) trees are informative structures in software or web applications,
each \atree node corresponds to a UI element on the screen.
}.
Such data enable end-to-end training of GUI agents, extending their capabilities from automating actions~\citep{cheng2024seeclick} to achieving full-process autonomy~\citep{zhang2024agentohana}. 

However, collecting such trajectories is far from trivial.
Existing \textit{task-driven} methods, which rely on humans or machines executing pre-defined tasks, face the following limitations:
human collection requires annotators to label entire trajectories and pre-define high-level tasks manually~\citep{li2024androidcontrol,lu2024weblinx},
making it both costly and labor-intensive.
Model-based synthesis also faces critical challenges:
(1) it heavily depends on pre-defined high-level tasks~\citep{lai2024autowebglm}, which not only limit the scalability of synthesized data but also constrain its diversity;
and 
(2) it struggles to ensure data quality, as errors in intermediate steps or mismatched task objectives can lead to incomplete or incoherent trajectories~\citep{murty2024bagel,patel2024large}.
Above mentioned issues pose a bottleneck for advancing GUI agents. 
Thus, effective trajectory construction methods are a clear desideratum for addressing these challenges.

In this paper, we present \ours, a pipeline for synthesizing high-quality and diverse GUI agent trajectories without involving human supervision or pre-defined tasks. 
Recognizing the limitations of the aforementioned \textit{task-driven} methods,
we draw inspiration from how humans learn to interact with GUI applications and adopt an \textit{interaction-driven} approach.
\ours begins by exploring the functionality of GUI environments through traversing interactive UI elements with actions (\textit{e.g.}, \texttt{CLICK}). 
This forms the basis for reverse task synthesis, 
where observed states and actions are retroactively transformed into low-level instructions. 
These low-level instructions are then derived into high-level instructions,
which can seed the collection of GUI trajectories.
By uncovering considerable functionalities,
reverse task synthesis facilitates the creation of meaningful and executable tasks.
Moreover, it naturally bridges the gap between abstract instructions and the dynamic nature of GUIs. 
Once synthesized tasks are converted into trajectories, we introduce a reward model to ensure data quality and effective utilization.

Experiments on two challenging online benchmarks, AndroidWorld and WebArena, demonstrate the effectiveness of \ours. 
It surpasses task-driven methods by a large margin, nearly doubling the performance from 9.82\% to 17.41\% on AndroidWorld. 
This highlights the high quality of trajectories synthesized by \ours and its great potential to transform general-purpose VLMs into specialized GUI agents.

Our primary contributions are as follows:
\begin{itemize}[itemsep=2pt,topsep=3pt,parsep=0pt]
\item By shifting from \textit{task-driven} approaches to \textit{interaction-driven} GUI agent data construction, we introduce reverse task synthesis to improve trajectory quality and diversity.
\item We propose a novel pipeline, \ours, capable of efficiently synthesizing high-quality trajectory data. Without human supervision, \ours supports end-to-end training of GUI agents across environments.
\item Extensive experiments across mobile and web tasks on dynamic benchmarks demonstrate the superior performance of \ours over a suite of strong baselines.
\end{itemize}

\section{Related Works}
\paragraph{Agents for Digital Automation.}
The recent proliferation of LLMs has significantly boosted researchers' interest in developing language agents~\citep{durante2024agent} to explore the digital world~\citep{feng2024far,wu2024oscopilot}. 
One line of work leverages the capabilities of fixed LLMs to create agents using methods like prompt engineering, model collaboration~\citep{wu2023autogen,sun2023corex,jia2024agentstore}, code or tool use~\citep{sun2024survey}, 
self-improvement~\citep{shinn2024reflexion,xu2024envisions,cheng2024vision}, 
or integration with world or agent models~\citep{hu2023language, jin2024mmtom, yang2023appagent}. 
Another line focuses on fine-tuning to augment models with agentic abilities, including (1) the ability to perceive the state of the computer, such as understanding screens~\citep{cheng2024seeclick,gou2024navigating,wu2024atlas} or application UI trees~\citep{OSWorld, zheng2024seeact}, (2) the ability to generate actions (click, type, scroll, \textit{etc}. \citealp{chen2024guicourse}), and (3) the flexibility to operate across diverse environments, including web~\citep{yao2022webshop,deng2023mindweb}, 
desktop~\citep{kapoor2024omniact,niu2024screenagent}, and mobile platforms~\citep{li2024androidcontrol,wang2024mobileagent}. Collectively, these efforts pave the way for digital automation, with agents engaging across a diverse digital landscape.

\begin{figure*}[ht]
  \centering
  \includegraphics[width=0.95\linewidth]{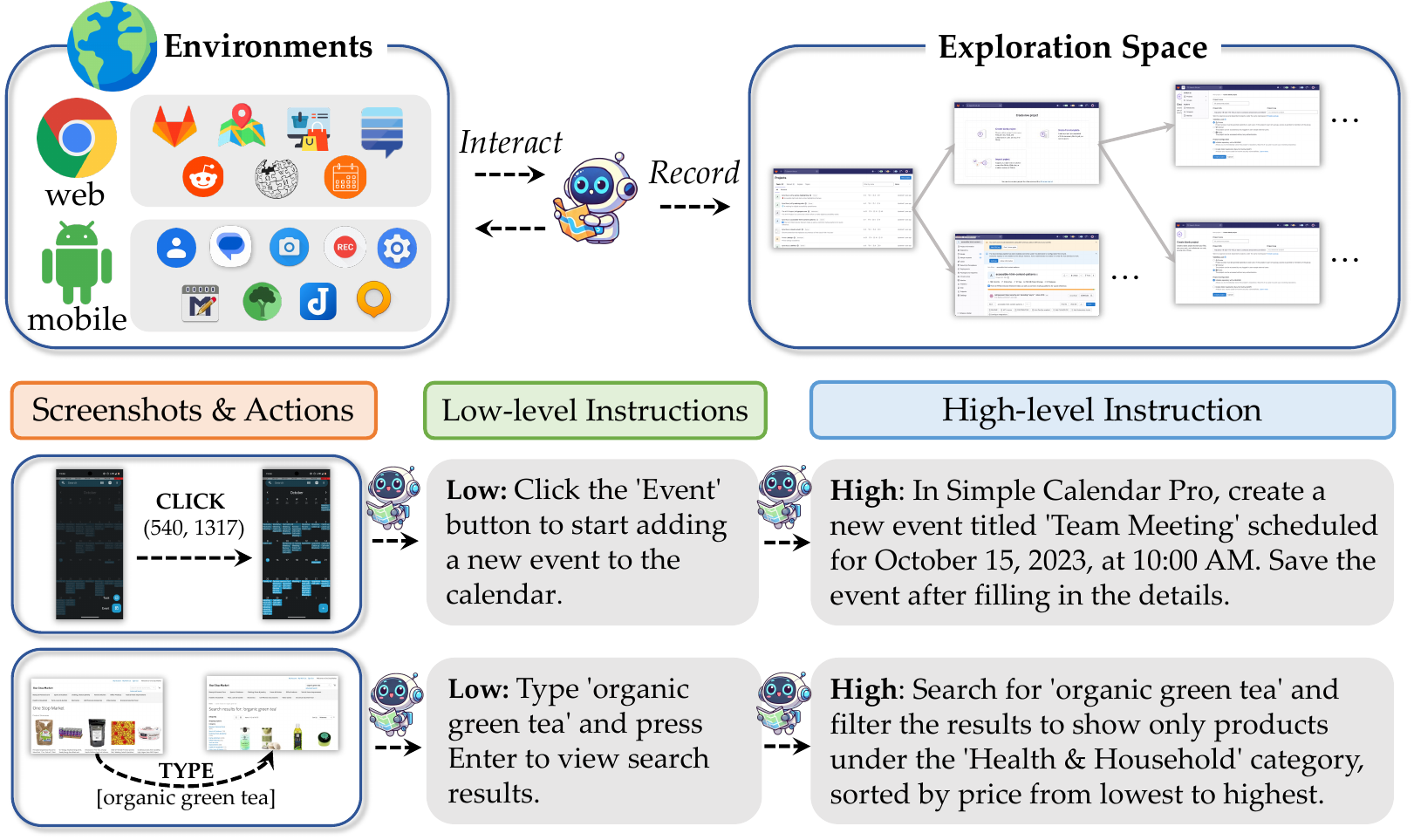}
    \vspace{-0.5em}
\caption{An overview of how we generate instruction data without relying on predefined tasks or human annotations.
\ours\ begins with a model-free, interaction-driven traversal in online environments (\textit{e.g.}, a web browser).
This process produces massive triples consisting of actions and their corresponding pre- and post-interaction screenshots.
Reverse task synthesis leverages these triples to generate low-level instructions and associates them with broader objectives to construct high-level instructions.
}
  \label{fig:architecture}
  \vspace{-0.75em}
\end{figure*}


\vspace{-0.5em}


\paragraph{Data for Building Computer Agents.}
High-quality GUI data are essential for bridging models from the symbolic world~\citep{xu2024symbol} to the digital world~\citep{wu2024oscopilot}, 
enabling the development of computer control agents.
Such data empower models to propose plans, execute appropriate actions, and autonomously navigate diverse environments~\citep{zeng2024agenttuning,pan2024autonomous}.
Rico~\citep{deka2017rico} first introduces sequential GUI data for mobile apps, while MiniWob~\citep{shi2017world} provides low-level keyboard and mouse actions for web-based tasks. 
Since then, several works have expanded the availability of such data for mobile~\citep{rawles2023aitw,zhang2024aitz,lu2024guiodyssey,chai2024amex}, 
web~\citep{zheran2018miniwobplus,lu2024weblinx,murty2024nnetscape},
and desktop~\citep{chen2024guicourse} applications.
To effectively build computer agents, 
the best approach is to use {trajectory data}, which should consist of sequences containing GUI information, both low-level and high-level instructions, as well as corresponding actions~\citep{li2024androidcontrol,zhang2024agentohana,zheng2024agentstudio}.
However, acquiring such trajectories poses significant challenges. 
First, existing datasets often lack essential components. 
Second, the reliance on manual curation makes data collection costly and inefficient.
Finally, current works are usually tailored to specific GUI (\textit{e.g.}, web-only), restricting their applicability in broader scenarios.

\section{\ours}

In this section, we present the pipeline of \ours, detailing the process from automated data collection to the construction of complete GUI agent trajectories. 

\begin{figure*}[ht]
  \centering
  \includegraphics[width=\linewidth]{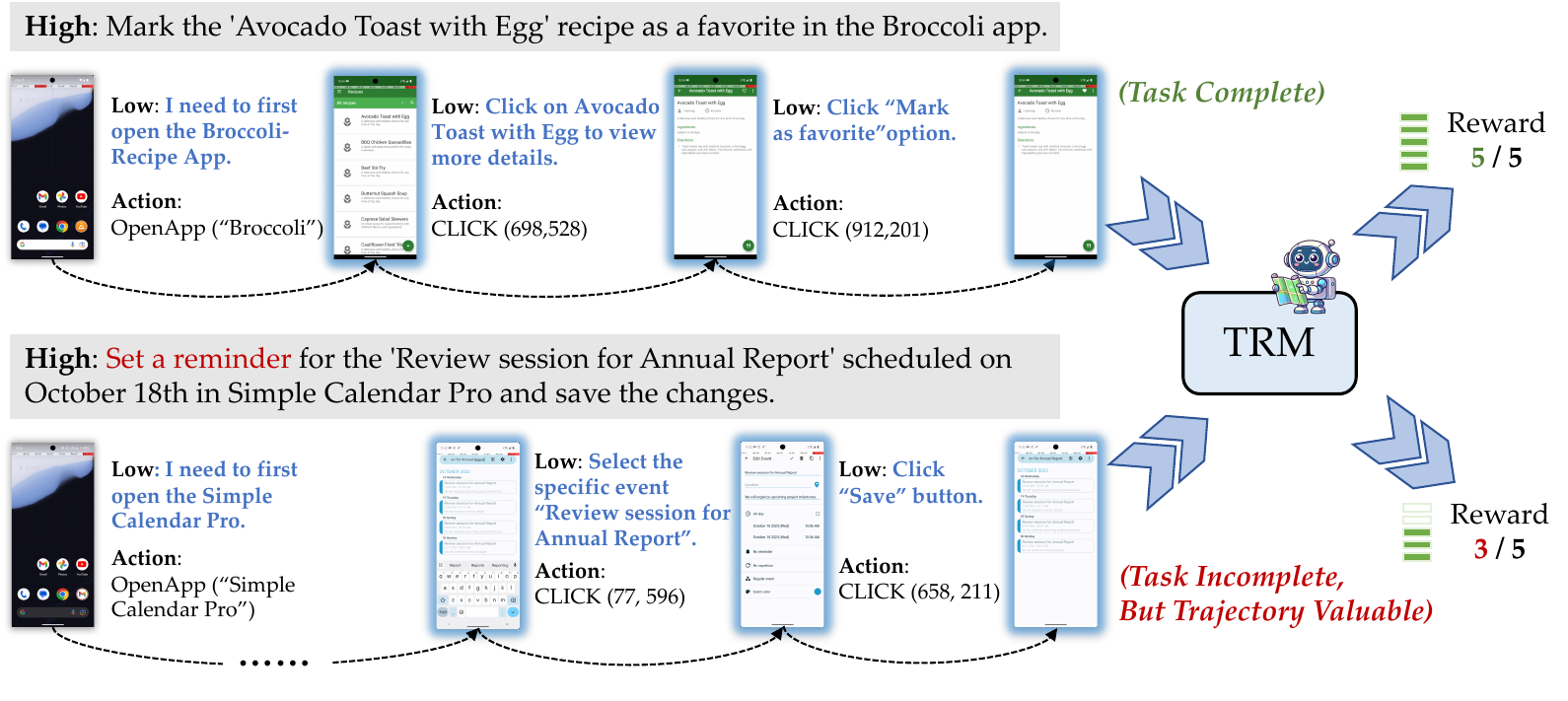}
\caption{An overview of collecting complete trajectories through exploring high-level instructions generated by reverse task synthesis. Low-level instructions and the last three states of the trajectory (indicated in {\color[HTML]{4472C4} light blue}) are used by the Trajectory Reward Model (TRM) to assign reward scores.}
  \label{fig:explore-reward}
  \vspace{-0.25em}
\end{figure*}

\subsection{Interaction-Driven Functional Discovery}

As illustrated in Figure~\ref{fig:architecture},
\ours\ begins with human-free
exploration in dynamic environments 
${\mathcal{E} = {\text{mobile}, \text{web}, \text{etc.}}}$,
systematically traversing interactive elements through actions ${a \in \mathcal{A}=\{\texttt{CLICK}, \texttt{TYPE}, \texttt{SCROLL}\}}$.
With the goal of constructing mobile and web agents,
this process is conducted in both the Android emulator \raisebox{0.095em}{\includegraphics[width=.35cm, valign=c]{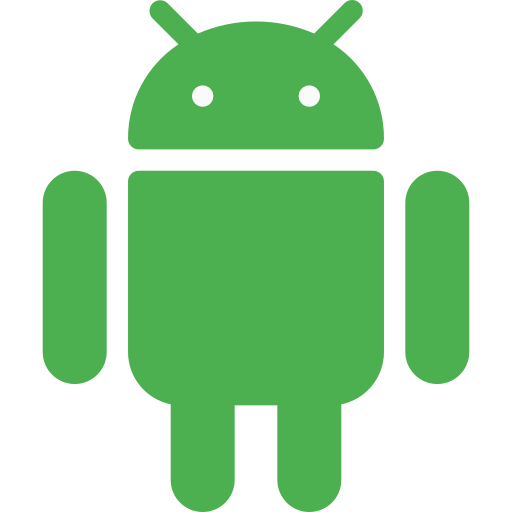}} and a chrome 
browser \raisebox{0.085em}{\includegraphics[width=.35cm, valign=c]{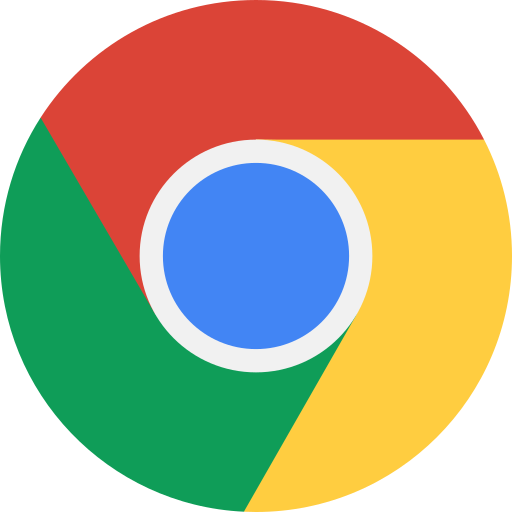}}~\footnote{We build dynamic environments on the basis of \citet{zhou2024webarena} and \citet{rawles2024androidworld}.}.
It to some extent mirrors human interaction with GUIs, uncovering potential functionalities without requiring pre-defined tasks.

The entire exploration phase is rule-based,
except when interacting with input fields, where \gpt\ is invoked to generate contextually appropriate contents.
At the end of this phase, 
massive triplets ${\langle s_{\text{pre}}, a, s_{\text{post}} \rangle}$ are collected,
where ${s_{\text{pre}}}$ and ${s_{\text{post}}}$ denote the pre- and post-action states (\textit{i.e.}, screenshots of the interface before and after the action),
and ${a}$ denotes the executed action.

\subsection{Reverse Task Synthesis}
Following the discovery, \ours leverages collected triplets ${\langle s_{\text{pre}}, a, s_{\text{post}} \rangle}$ to construct meaningful task instructions.
This process involves generating low-level tasks using an annotation model~(\textit{e.g.}, \gpt) and subsequently transforming them into high-level tasks.
The annotation model $\mathcal{M}$ transforms each triplet ${\langle s_{\text{pre}}, a, s_{\text{post}} \rangle \in \mathcal{T}}$ into a specific low-level task instruction:
\[
f_{\text{low}}: \langle s_{\text{pre}}, a, s_{\text{post}} \rangle \xrightarrow{\mathcal{M}} \tau_{\text{low}}.
\]
Here, ${\tau_{\text{low}}}$ represents an atomic, executable operation derived from the observed state transition caused by the action ${a}$. 
For example, if the action ${a = \texttt{CLICK}}$ reveals a dropdown menu, the corresponding task might be ``click the dropdown to display options.'' 
The annotation model integrates visual, contextual, and action semantics to ensure that ${\tau_{\text{low}}}$ aligns with the functions of $\mathcal{E}$.

Building on the synthesized low-level tasks, \ours\ constructs high-level tasks by associating each low-level task ${\tau_{\text{low}}}$ with broader objectives that could plausibly encompass it. 
This process, performed by the annotation model $\mathcal{M}$, maps individual low-level steps to high-level tasks by leveraging contextual information and domain knowledge:
\[
f_{\text{high}}: \tau_{\text{low}} \xrightarrow{\mathcal{M}} \tau_{\text{high}}.
\]
Here, ${\tau_{\text{high}}}$ represents a goal-oriented instruction that contextualizes the low-level operation within a larger user intent. For instance, a low-level task such as ``click the dropdown to display options'' might be linked to a high-level task like “configure application settings,” as the dropdown interaction is often a prerequisite for such configurations. 
Details and prompts for transforming triples into high-level instructions are provided in Appendix~\ref{app:rev}.

After this reverse task synthesis process, \ours\ generates a diverse set of high-level instructions ${\mathcal{T} = \{\tau_1, \tau_2, \dots, \tau_N}\}$ that are aligned with dynamic environments and semantically rich. 
This entire process is completed without any human intervention. 

Subsequently, these synthetic instructions ${\mathcal{T}}$ are executed in environment ${\mathcal{E}}$ by a model like \gpt,
producing a complete set of trajectories, denoted as ${\mathcal{G} = \{g_1, g_2, \dots, g_N}\}$.



\subsection{Trajectory Reward Model}

Considering the potential limitations of a model’s agentic ability, errors or incomplete steps may arise when using high-level instructions to explore and generate trajectories.
To address this, we incorporate a Trajectory Reward Model (TRM) to ensure the quality and utility of trajectories synthesized by \ours,
as illustrated in Figure~\ref{fig:explore-reward}.
Previous methods commonly rely on labeler functions~\citep[][\emph{inter alia}]{he2024openwebvoyager,murty2024nnetscape},
which discard trajectories deemed incomplete directly~\citep{pan2024autonomous}. 
However, even incomplete trajectories often contain a valuable exploration of the GUI environment.
Given their large proportion of the data, discarding them wastes critical opportunities to enhance the model's agentic capabilities.
Thus, diverging from binary evaluation, we leverage the characteristics of trajectory.
Built upon \gpt, TRM aims to perform a graded evaluation with a reward score  $R \in [1, 5]$  to assist in sampling for training. Reward modeling focuses on the following features:
\vspace{-0.25em}
\begin{itemize}[itemsep=2pt,topsep=3pt,parsep=0pt]
\item \textbf{Completion}: Measures the extent to which the trajectory successfully fulfills the instructed task, considering completeness and proper handling of interactions.
\item \textbf{Coherence}: Evaluates whether the trajectory follows a logical sequence of actions toward achieving the high-level task, avoiding redundant or irrelevant steps.

\end{itemize}

\begin{algorithm}[ht]
\caption{Reward-Based Trajectory Sampling}
\begin{algorithmic}[1]
\Require 
    Trajectory set \( \mathcal{G} = \{ g_1, g_2, \dots, g_N \} \), where \( g_i = \{ s_{i,1}, l_{i,1}, s_{i,2}, \dots, s_{i,K_i} \} \) represents a trajectory with \( K_i \) steps, including states \( s_{i,j} \) and low-level instructions \( l_{i,j} \).
    Reward model \( \mathcal{RM} \).

\Ensure 
    Trajectories are sampled for training according to their rewards.

\For{each trajectory \( g_i \in \mathcal{G} \)}
    \State \textbf{Initialize} trajectory reward \( R_i \gets 0 \)
    \State \textbf{Extract} low-level instructions \( \mathcal{L}_i = \{ l_{i,1}, l_{i,2}, \dots, l_{i,K_i} \} \)
    \State \textbf{Extract} the last three states \( S_{\text{last}} = \{ s_{i,K_i-2}, s_{i,K_i-1}, s_{i,K_i} \} \)
    \State \textbf{Compute} trajectory reward: \( R_i = \mathcal{RM}(\mathcal{L}_i, S_{\text{last}}) \)
\EndFor

\For{each training iteration}
    \State \textbf{Compute} sampling probabilities \( P(g_i) = R_i\ /\ \left( \sum_{k=1}^{N} R_k \right) \) for all \( g_i \)
    \State \textbf{Sample} a trajectory \( g_i \) based on \( P(g_i) \) for each training step
\EndFor
\end{algorithmic}
\label{alg:trm}
\end{algorithm}

The whole process is shown in Algorithm~\ref{alg:trm}.
By leveraging TRM, \ours ensures that synthesized trajectories are utilized effectively, allowing the training process to benefit from both high-quality data and diverse task scenarios.
Further details, along with consistency analyses between human annotators and different VLM backbones are shown in Appendix~\ref{app:trm}.
\section{Experiments}

\begin{table*}[ht]
\centering
\resizebox{0.98\linewidth}{!}{
\begin{tabular}{llcccccc}
\toprule
\multirow{2}{*}{\textbf{Base Model}} & \multirow{2}{*}{\textbf{Strategies}} & \multirow{2}{*}{\textbf{AndroidWorld}} & \multicolumn{2}{c}{\textbf{AndroidControl-High}} & \multicolumn{2}{c}{\textbf{AndroidControl-Low}} \\
 &  &  & \textbf{SR} & \textbf{Type} & \textbf{SR} & \textbf{Type} \\
\midrule
\texttt{GPT-4o} & Zero-Shot (M3A) & {23.70} & 53.04 & 69.14 & 69.59 & 80.27 \\
\midrule
\multirow{5}{*}{\begin{tabular}[c]{@{}c@{}}InternVL2-4B\end{tabular}} & {Zero-Shot} & 0.00 & 16.62 & 39.96 & 33.69 & 60.65 \\
 & {Task-Driven} & 4.02 & 27.37 & 47.08 & 66.48 & 90.37 \\
 & Task-Driven w. Self Instruct & 7.14 & 24.95 & 44.27 & 66.70 & 90.79 \\
\cmidrule(lr){2-7}
 & {\ours} & \textbf{{15.18}} & \textbf{{33.39}} & \textbf{{56.20}} & \textbf{{73.38}} & \textbf{{91.32}} \\
\midrule
\multirow{5}{*}{\begin{tabular}[c]{@{}c@{}}InternVL2-8B\end{tabular}} & {Zero-Shot} & 2.23 & 17.89 & 38.22 & 47.69 & 66.67 \\
 & {Task-Driven} & 4.46 & 23.79 & 43.94 & 64.43 & 89.83 \\
 & Task-Driven w. Self Instruct & 5.36 & 23.43 & 44.43 & 64.69 & 89.85 \\
\cmidrule(lr){2-7}
 & {\ours} & \textbf{{16.96}} & \textbf{{35.77}} & \textbf{{64.57}} & \textbf{71.37} & \textbf{91.27} \\
\midrule
\multirow{5}{*}{\begin{tabular}[c]{@{}c@{}}Qwen2-VL-7B\end{tabular}} & 
{Zero-Shot} & 0.89 & 28.92 & 61.39 & 46.37 & 72.78 \\
 & {Task-Driven} & 6.25 & 38.84 & 58.08 & 71.33 & 88.71 \\
 & Task-Driven w. Self Instruct & 9.82 & 39.36 & 58.28 & 71.51 & 89.73 \\
\cmidrule(lr){2-7}
 & {\ours} & \textbf{17.41} & \textbf{44.54} & \textbf{66.15} & \textbf{74.17} & \textbf{90.72} \\
\bottomrule
\end{tabular}
}
\caption{Evaluations on AndroidControl and AndroidWorld. {SR} represents the task success rate.
{Type} measures the exact match score between the predicted action types (\textit{e.g.}, \texttt{CLICK}, \texttt{SCROLL}) and the ground truth.}
\label{tab:android_results}
\vspace{-0.75em}
\end{table*}


\vspace{-0.25em}
\subsection{Experimental Settings}
\paragraph{Evauation Benchmarks.} 
For mobile tasks, we select (1) AndroidControl~\citep{li2024androidcontrol}, which evaluates the ability of GUI agents to perform both low- and high-level tasks, 
and (2) AndroidWorld~\citep{rawles2024androidworld}, a challenging online benchmark running in Android emulators, to demonstrate the practicability of our agents in solving human daily tasks.
Regarding web tasks,
More information about the benchmark settings and evaluation details are presented in Appendix~\ref{app:bench}.

\vspace{-0.25em}
\paragraph{Model Settings.} 
We primarily use \gpt~\citep{hurst2024gpt} for reverse task synthesis and reward modeling. As for the backbone models used to construct agents, we consider (1) InternVL2-4B/8B~\citep{chen2023internvl}, which is trained without GUI data, 
and (2) Qwen2-VL-7B-Instruct~\citep{Qwen2VL}, which claims to possess certain agentic capabilities
to conduct thorough and comparative experiments.
All training is performed as VLM full fine-tuning on interconnected clusters of 8 $\times$ A100 80GB GPUs,
with detailed training settings provided in Appendix~\ref{app:exp_details} and prompt settings in Appendix~\ref{app:training_details}.


\vspace{-0.25em}
\subsection{Baseline Construction and Training}
\paragraph{Baselines.}
As a pioneering study in synthesizing GUI agent data,
we design the following baselines to demonstrate the superiority of trajectories obtained through \ours. All settings uniformly accept \atree and screenshots as inputs.
\begin{itemize}[itemsep=2pt,topsep=3pt,parsep=0pt]
    \item \textbf{Zero-Shot}: This baseline leverages CoT~\citep{wei2022chain} prompting to guide the model in perceiving environments and taking actions. 
    For AndroidWorld tasks, we follow \citet{rawles2024androidworld} to adopt M3A agent setup with multimodal input for this setting.
    \item \textbf{Task-Driven}: We build this baseline to compare with the common approach for agent data synthesis~\citep[][\emph{inter alia}]{lai2024autowebglm}.
    Given the initial screenshots of the app/web page and task examples, use \gpt to generate high-level instructions and explore the environment to collect trajectories. These trajectories are then used for training.  
    \item \textbf{Self-Instructions}: Building upon the task-driven baseline, 
    this approach employs \gpt to perform self-instruction~\citep{wang2023selfinstruct}, 
    generating additional high-level tasks for exploration and trajectory collection. Together with the previously collected trajectories, they are then used for training.  
\end{itemize}

Details of the baseline construction are provided in Appendix~\ref{sec:baseline}. All these data and resources will be made public to accelerate future research.

\vspace{-0.25em}
\paragraph{Trajectory Training.} Training GUI Agents based on VLMs using trajectory data is essentially a supervised fine-tuning (SFT) process. Nevertheless, we devise two training objectives to maximize the utility of synthesized trajectories:
\begin{itemize}[itemsep=2pt,topsep=3pt,parsep=0pt]
    \item \textbf{Planning Training.} This objective aims to enhance agents' planning ability.
    For each trajectory $g_i \in \mathcal{G}$, 
    given multimodal input $s$, high-level instruction $h_i$, and history context $c$, the agent $\theta$ predict the low-level instruction $\ell$ and the corresponding action $a$. 
    \begin{equation}
        \hspace*{-1em} 
        \mathcal{L}_{1} = - \sum\limits_{{t_i \in \mathcal{T}}} \log \Big( p_{\theta}(\ell \mid s, h_i, c) \cdot p_\theta(a \mid s, h_i, c, \ell) \Big)
    \end{equation}
    \item \textbf{Action Training.} This objective strengthens the agent's ability to execute appropriate actions based on the low-level instruction $\ell$.
    given $s$, $h_i$, $c$, the agent predicts the action $a$. 
    \begin{equation}
        \hspace*{-1em} 
        \mathcal{L}_{2} = - \sum\limits_{{t_i \in \mathcal{T}}}\log p_\theta(a \mid s, c, \ell)
        \vspace{-0.5em}
    \end{equation}
\end{itemize}

After trajectory training, 
agents will generate ReAct-style~\citep{yao2023react} outputs,
with their step-by-step thoughts recorded in the history.
To ensure a fair comparison, both the Task-Driven baseline and \ours\ use 1K trajectories for training,
while the Self-Instruction baseline uses 1.5K trajectories,
with an average trajectory length of 6.4 steps.

\begin{table*}[ht]
\centering
\resizebox{0.95\linewidth}{!}{
\begin{tabular}{clcccccc}
\toprule
{\textbf{Model}} & {\textbf{Strategies}} & \textbf{Shopping} & \textbf{CMS} & \textbf{Reddit} & \textbf{Gitlab} & \textbf{Maps} & \textbf{Overall} \\
\midrule
\texttt{GPT-4o} & Zero-Shot & 14.28 & 21.05 & 6.25 & 14.29 & 20.00 & 16.25 \\
\midrule
\multirow{5}{*}{\begin{tabular}[c]{@{}c@{}}InternVL2-4B\end{tabular}} & {Zero-Shot} & 0.00 & 0.00 & 0.00 & 0.00 & 0.00 & 0.00 \\
 & {Task-Driven} & 5.36 & 1.76 & 0.00 & \textbf{9.52} & 5.00 & 4.98 \\
 & Task-Driven w. Self-Instruct & 5.36 & 3.51 & 0.00 & \textbf{9.52} & \textbf{7.50} & 5.81 \\
\cmidrule(lr){2-8}
 & {\ours} & \textbf{10.71} & \textbf{7.02} & \textbf{3.13} & 7.94 & \textbf{7.50} & \textbf{7.88} \\
\midrule
\multirow{5}{*}{\begin{tabular}[c]{@{}c@{}}InternVL2-8B\end{tabular}} & {Zero-Shot} & 0.00 & 0.00 & 0.00 & 0.00 & 0.00 & 0.00 \\
 & {Task-Driven} & 3.57 & 7.02 & 0.00 & 6.35 & 2.50 & 4.56 \\
 & Task-Driven w. Self-Instruct & \textbf{8.93} & 10.53 & 6.25 & \textbf{7.94} & 0.00 & 7.05 \\
\cmidrule(lr){2-8}
 & {\ours} & 7.14 & \textbf{15.79} & \textbf{9.34} & 6.35 & \textbf{10.00} & \textbf{9.96} \\
\midrule
\multirow{5}{*}{\begin{tabular}[c]{@{}c@{}}Qwen2-VL-7B\end{tabular}} & {Zero-Shot} & \textbf{12.50} & 7.02 & 6.25 & 6.35 & 5.00 & 7.47 \\
 & {Task-Driven} & 8.93 & 7.02 & 6.25 & 6.35 & 5.00 & 7.05 \\
 & Task-Driven w. Self-Instruct & 8.93 & 1.76 & 3.13 & 4.84 & \textbf{7.50} & 5.39 \\
\cmidrule(lr){2-8}
 & {\ours} & 7.14 & \textbf{8.77} & \textbf{15.63} & \textbf{15.87} & 5.00 & \textbf{{10.79}} \\
\bottomrule
\end{tabular}
}
\caption{Evaluations on WebArena with success rate reported.}
\label{tab:web_tasks}
\vspace{-0.5em}
\end{table*}

\subsection{Main Results}

\paragraph{AndroidWorld.} 
To prove the effectiveness of \ours under dynamic environment,
we evaluate it on AndroidWorld~\citep{rawles2024androidworld} that leverages a Pixel 6 phone simulator as testbed.
As shown in Table~\ref{tab:android_results},
\ours significantly narrows the performance gap between open-source agents and the SOTA \gpt-based M3A agent.
Compared to task-driven methods, 
training with \ours\ achieves performance improvements that are often double those of the baselines.
Even self-instruct baseline utilize 1.5 $\times$ the amount of data compared to \ours, they fail to match the quality of data generated by \ours.
underscoring the importance of using high-quality trajectory data in online settings.

Beyond improvements in planning and action,
some gains also stem from \ours’s ability to cover subtle yet critical app functionalities during the reverse task synthesis process.
These functionalities, 
often overlooked by task-driven methods, are essential for completing intricate tasks.

\vspace{-0.25em}
\paragraph{AndroidControl.} 
We then evaluate \ours\ on AndroidControl~\citep{li2024androidcontrol}.
Out of the 833 apps covered by AndroidControl, only 20 have been directly encountered during data synthesis,
making this evaluation a test of \ours’s out-of-distribution (OOD) performance.
In the high-level setting, 
the agent is required to autonomously plan and execute actions to complete a given task.
For the low-level setting, 
agents will follow human instructions and only need to determine the next step.
As shown in Table~\ref{tab:android_results},
\ours consistently improves both action and planning abilities across various backbones.
Compared to \gpt, \ours achieves substantial gains, especially in the low-level setting where it consistently outperforms. 
While maintaining an edge over other task-driven trajectory synthesis methods, 
\ours excels particularly in the high-level setting.
This validates that exploration-first task construction produces more meaningful and logically coherent tasks. 
Additionally, it highlights \ours’s generalization ability to unseen OOD scenarios compared to task-driven approaches.

\vspace{-0.25em}
\paragraph{WebArena.}
We choose WebArena~\citep{zhou2024webarena}, 
a highly challenging benchmark running on functional websites to evaluate \ours on web environments.
We follow similar baseline settings as in mobile tasks.
Results in Table~\ref{tab:web_tasks} show that training with \ours\ data generally leads to notable performance improvements.
For InternVL2-4B and 8B that can hardly generate outputs in correct formats under zero-shot settings, 
\ours\ enables a remarkable leap in performance after training.
For Qwen2-VL-7B, which has already been trained on GUI agent data, further training with \ours\ results in substantial performance gains.
Notable edges over task-driven baselines highlight that, in web environments rich with interactive elements,
reverse task synthesis can derive more meaningful explorations.

\section{Analysis}
\label{sec:analysis}

\subsection{How Diverse is Our Synthesized Data?}
\label{sec:diversity_analysis}



Ensuring the diversity of synthetic data is crucial for effective model training. Traditional approaches that rely on pre-defined high-level tasks are inherently constrained, as it is practically impossible to enumerate and cover the full spectrum of potential interactions within a complex environment. In contrast, \ours employs an exploration-driven method that naturally adapts to the environment by interacting with diverse interface elements, systematically uncovering a broader range of functional capabilities. 

\begin{figure}[t]
    \centering
    \begin{subfigure}[b]{0.48\textwidth}
        \centering
        \includegraphics[width=\textwidth]{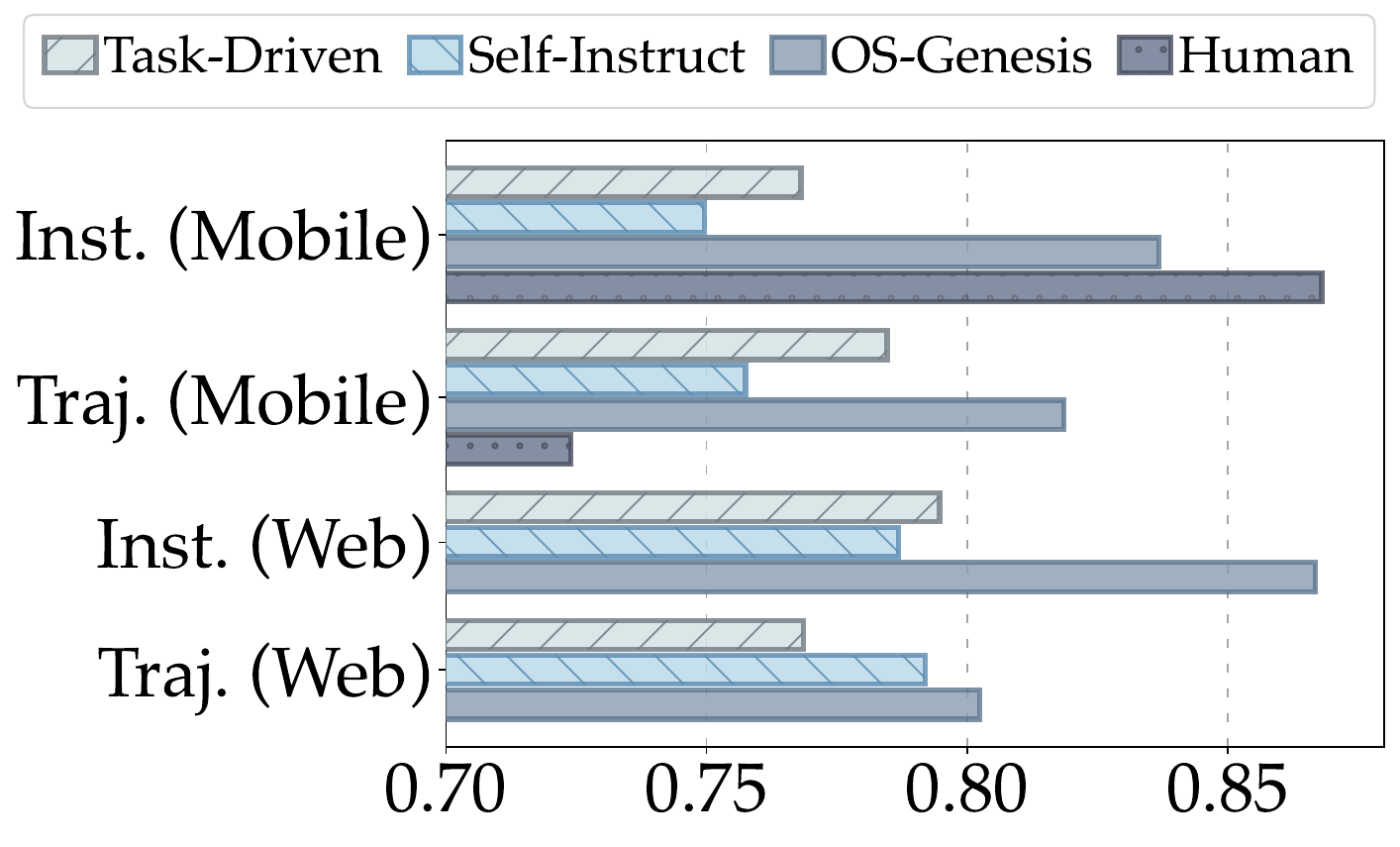}
    \end{subfigure}
    \caption{Comparison of instruction diversity and trajectory diversity between different synthetic data and human data, measured by average cosine distance.}
    \label{fig:data_diversity_analysis}
    \vspace{-0.75em}
\end{figure}

To validate the effectiveness of our method in generating more diverse data, we examine both instruction diversity and trajectory diversity.
We begin by analyzing the variety of generated instructions. Using Sentence-BERT~\citep{reimers2019sentencebert}, we embed each instruction and compute the average cosine distance among these embeddings. As illustrated in Figure~\ref{fig:data_diversity_analysis}, \ours achieves the greatest average distance across both mobile and web environments among different synthetic data, indicating a broader range of task types beyond those pre-defined at the outset. We then apply the same approach to the low-level actions taken in the generated trajectories. \ours demonstrates the highest trajectory diversity, suggesting that our interaction-driven strategy more thoroughly exploits the available operations within different environments. Additional visualizations and details are provided in Appendix~\ref{app:diversity_details}.

Interestingly, while human-annotated data displays high instruction diversity, it shows low trajectory diversity. This suggests that while humans can imagine various instructions, they tend to rely on a narrower set of familiar, well-practiced actions for execution. In contrast, \ours achieves high diversity in both instructions and trajectories, enabling a more comprehensive exploration of the environment.

\subsection{How TRM Impacts Performance?}

We introduce a Trajectory Reward Model (TRM) for data quality control and exploitation,
substituting traditional labeler filtering methods~\citep{he2024openwebvoyager,murty2024nnetscape}.
To analyze its impact and for ablation purposes,
we include additional settings for comparison:
(1) training without an RM, where all synthesized data is treated equally during training,
and 
(2) using a labeler, similar to previous approaches where only complete trajectories are retained for training.

\begin{figure}[ht]
    \centering
    \includegraphics[width=1.03\linewidth]{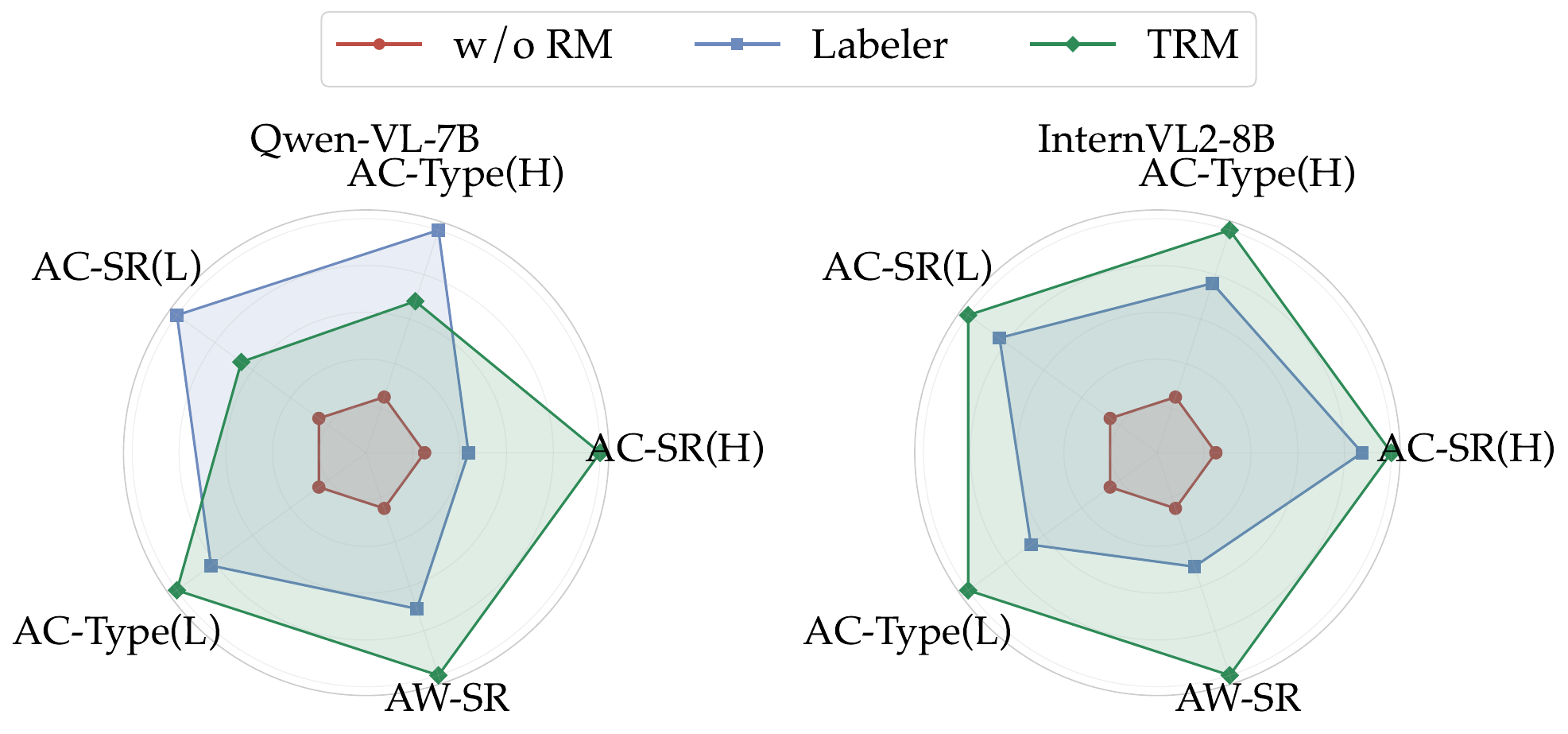}
    \caption{Comparison of different reward modeling strategies.
    }
    \label{fig:rm-ablation}
\end{figure}

As shown in Figure~\ref{fig:rm-ablation}, 
the relative performance across different reward strategies demonstrates the effectiveness of TRM, notably in enhancing high-level capabilities (\textit{e.g.}, AndroidControl-High and AndroidWorld).
While using a labeler provides slight gains in high-level tasks, it comes at the cost of reduced performance in low-level tasks.
For low-level scenarios, since \ours data—even individual steps—is inherently more meaningful and of good quality, all training strategies yield consistent improvements.

\subsection{How Scaling Trajectory Data Improves Agentic Ability?}
We investigate the impact of data scale on building GUI agents. 
To explore this, 
we partition the data synthesized by \ours into subsets, 
ranging from small-scale trajectories to those exceeding the size used in main experiments.
Using AndroidWorld as our testbed, we focus on two primary questions:
(1) How does performance improve as the data scale increases?
(2) Does performance saturate at higher data scales?

\begin{figure}[ht]
    \centering
    \begin{subfigure}[b]{0.48\textwidth}
        \centering
        \includegraphics[width=\textwidth]{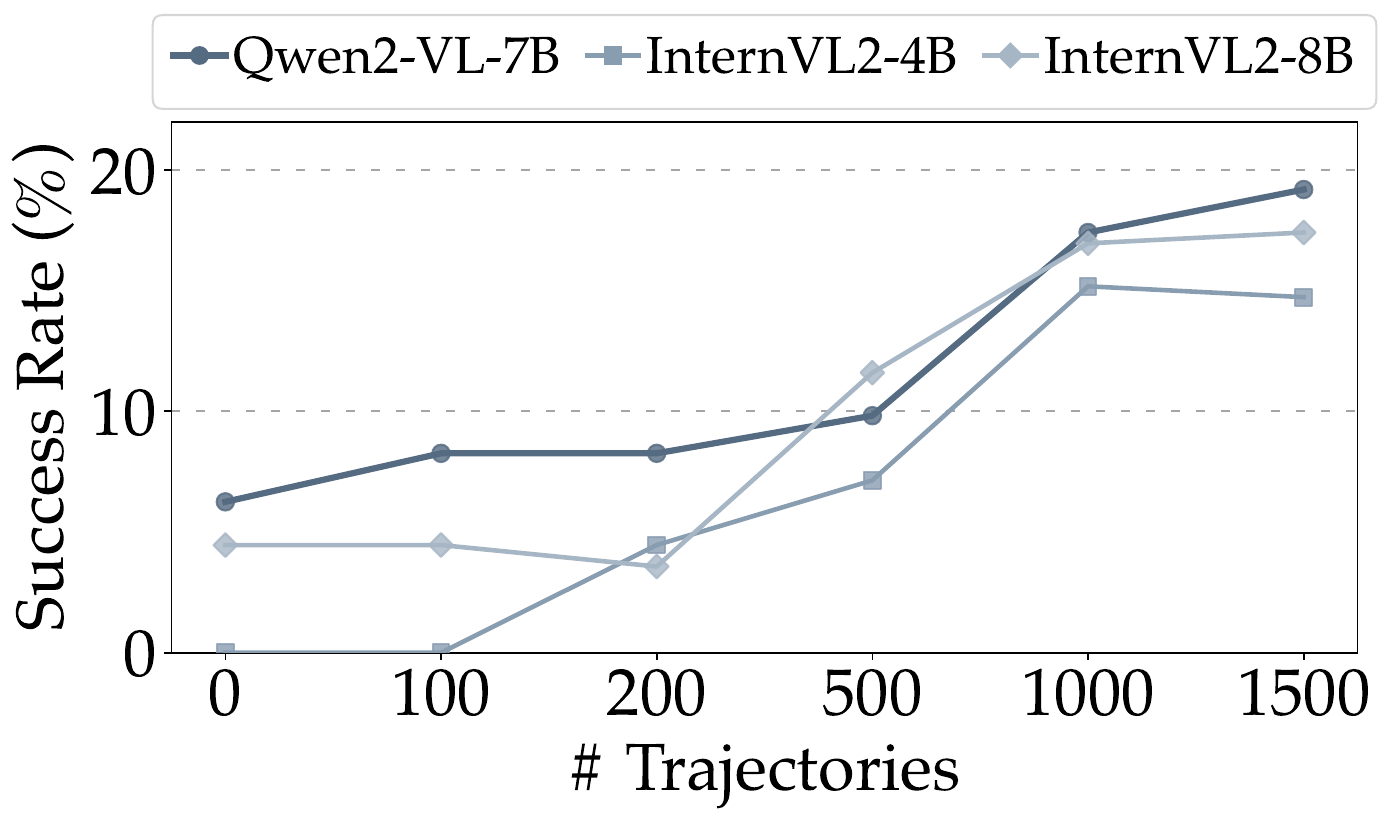}
    \end{subfigure}
    \caption{Performance of GUI agents trained on datasets of varying scales.}
    \label{fig:scaling}
    \vspace{-0.75em}
\end{figure}

As shown in Figure~\ref{fig:scaling}, task performance generally improves as the number of trajectories increases, while saturation emerges at larger data scales. We attribute this saturation to two key factors:
(1) The inherent capacity limitations of VLMs, and
(2) The constraints imposed by the exploration space and the ability of \gpt to materialize high-level instructions into complete trajectories effectively.
Overall, the data provided by \ours adequately supports the effective transformation of VLMs into GUI agents.

\subsection{How Far are We from Human Data?}
We analyze the gaps between \ours and human data in two key aspects:
(1) high-level instructions synthesized through \ours v.s. human-written instructions, and
(2) trajectories from \ours v.s. human-annotated trajectories.

\begin{figure}[ht]
\vspace{-0.25em}
    \centering
    \begin{subfigure}[b]{0.41\textwidth}
        \centering
        \includegraphics[width=\textwidth]{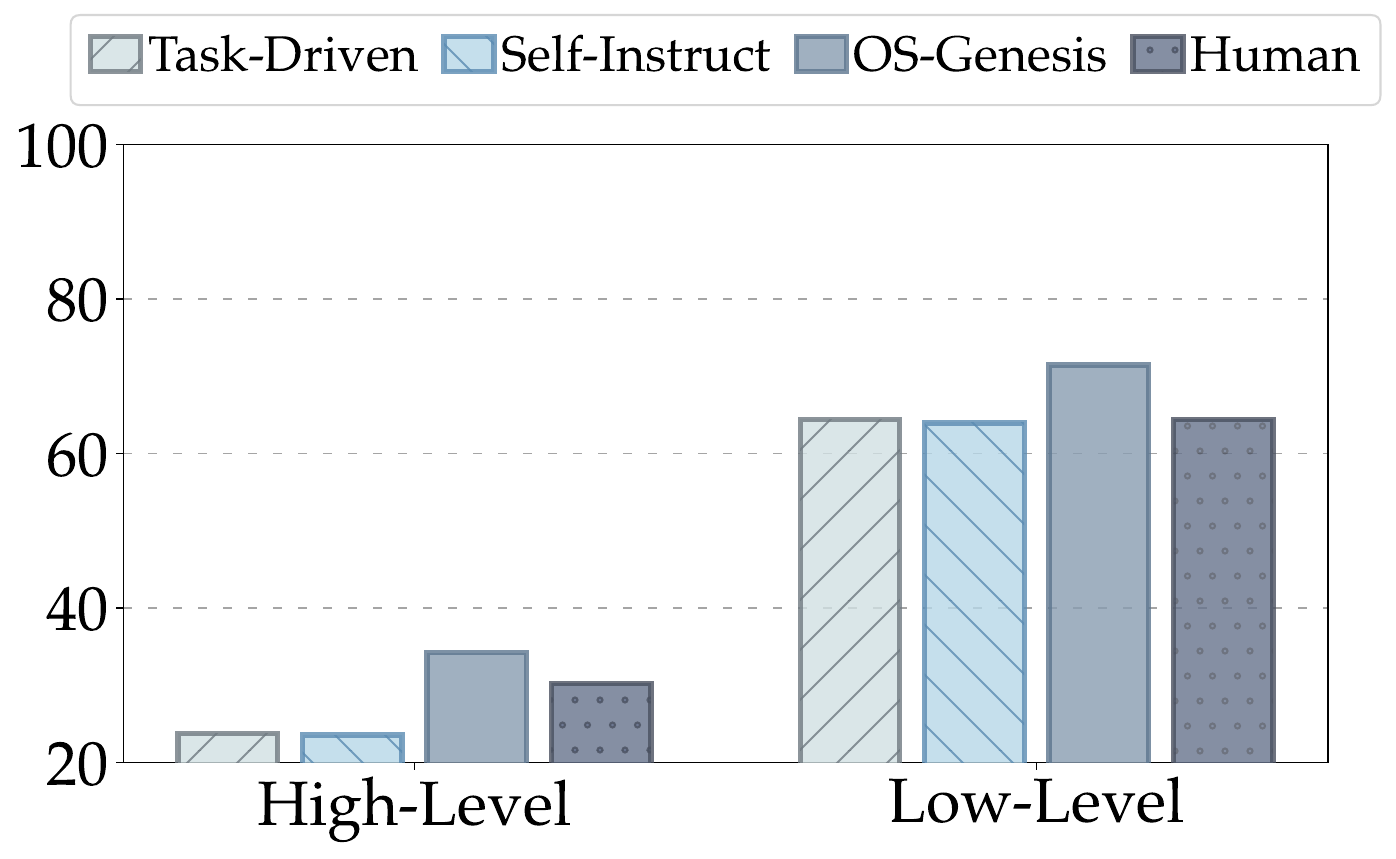}
        \subcaption{InternVL2-8B}
    \end{subfigure}
        \vspace{0.75em} 
    \begin{subfigure}[b]{0.41\textwidth}
        \centering
        \includegraphics[width=\textwidth]{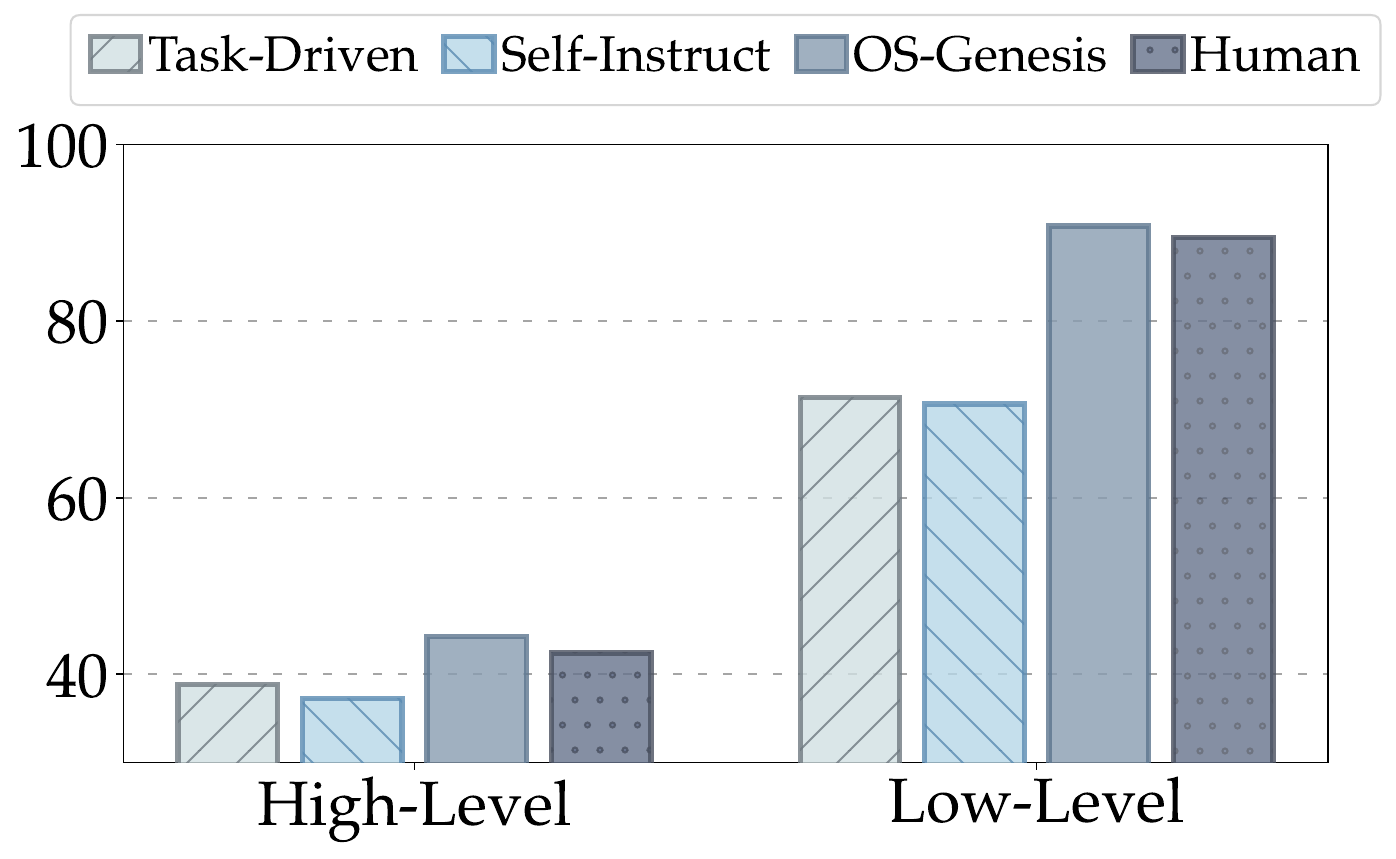}
        \subcaption{Qwen2-VL-7B-Instruct}
    \end{subfigure}
     \vspace{-0.5em}
    \caption{Comparison of training effectiveness between trajectories constructed from human-written and \ours high-level instructions.}
    \label{fig:data_diff_high_analysis}
    \vspace{-0.75em}
\end{figure}

\paragraph{High-Level Instructions.}
We first compare high-level instructions written by humans with those generated through reverse task synthesis by \ours. 
Based on the available apps in AndroidWorld,
we match 500 human-written tasks from the AndroidControl training set and use \gpt for exploration. 
The collected trajectories are then used to train agents based on InternVL2-8B and Qwen2-VL-7B. For comparison, an equal amount of \ours and baseline data is used for training. The results are presented in Figure~\ref{fig:data_diff_high_analysis}.

As observed, even when high-level instructions are written by human,
their performance falls short compared to \ours's instructions. 
This can be attributed to two main factors:
(1) Pre-defined tasks sometimes fail to align with the dynamic environment, and
(2) Models may introduce errors when interpreting the intentions of human annotators.
In contrast, \ours generates data in a progressive way, grounded in low-level interactions, which makes it inherently more suitable for unsupervised exploration and adaptation.

\paragraph{Trajectories.}
Here, we investigate the gaps between complete \ours trajectories and human demonstrations in GUI agent training.
We select 1K crowdsourced trajectories from AndroidControl training set for comparison.
\begin{figure}[ht]
\vspace{-0.25em}
    \centering
    \begin{subfigure}[b]{0.41\textwidth}
        \centering
        \includegraphics[width=\textwidth]{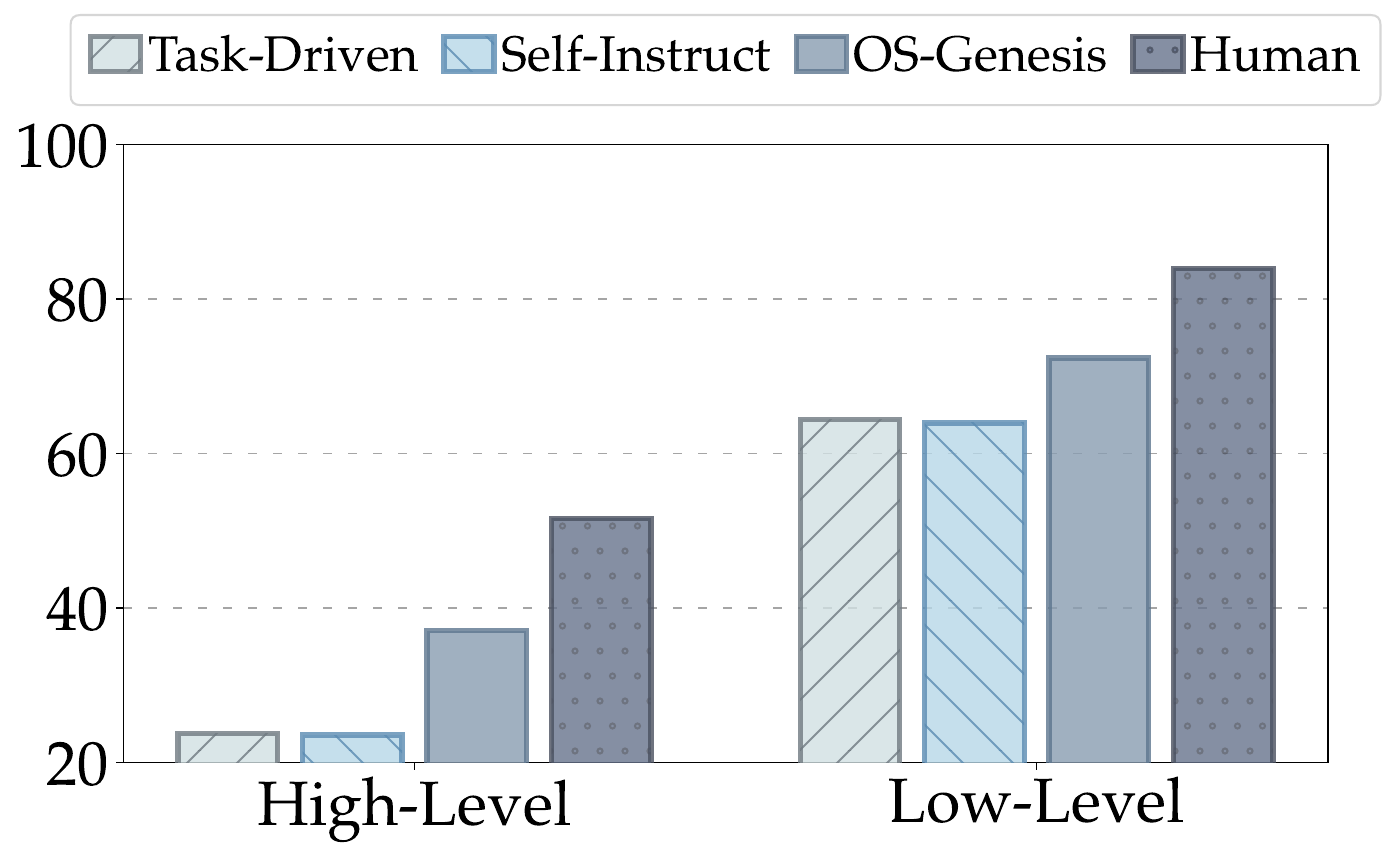}
        \subcaption{InternVL2-8B}
    \end{subfigure}
        \vspace{0.75em} 
    \begin{subfigure}[b]{0.41\textwidth}
        \centering
        \includegraphics[width=\textwidth]{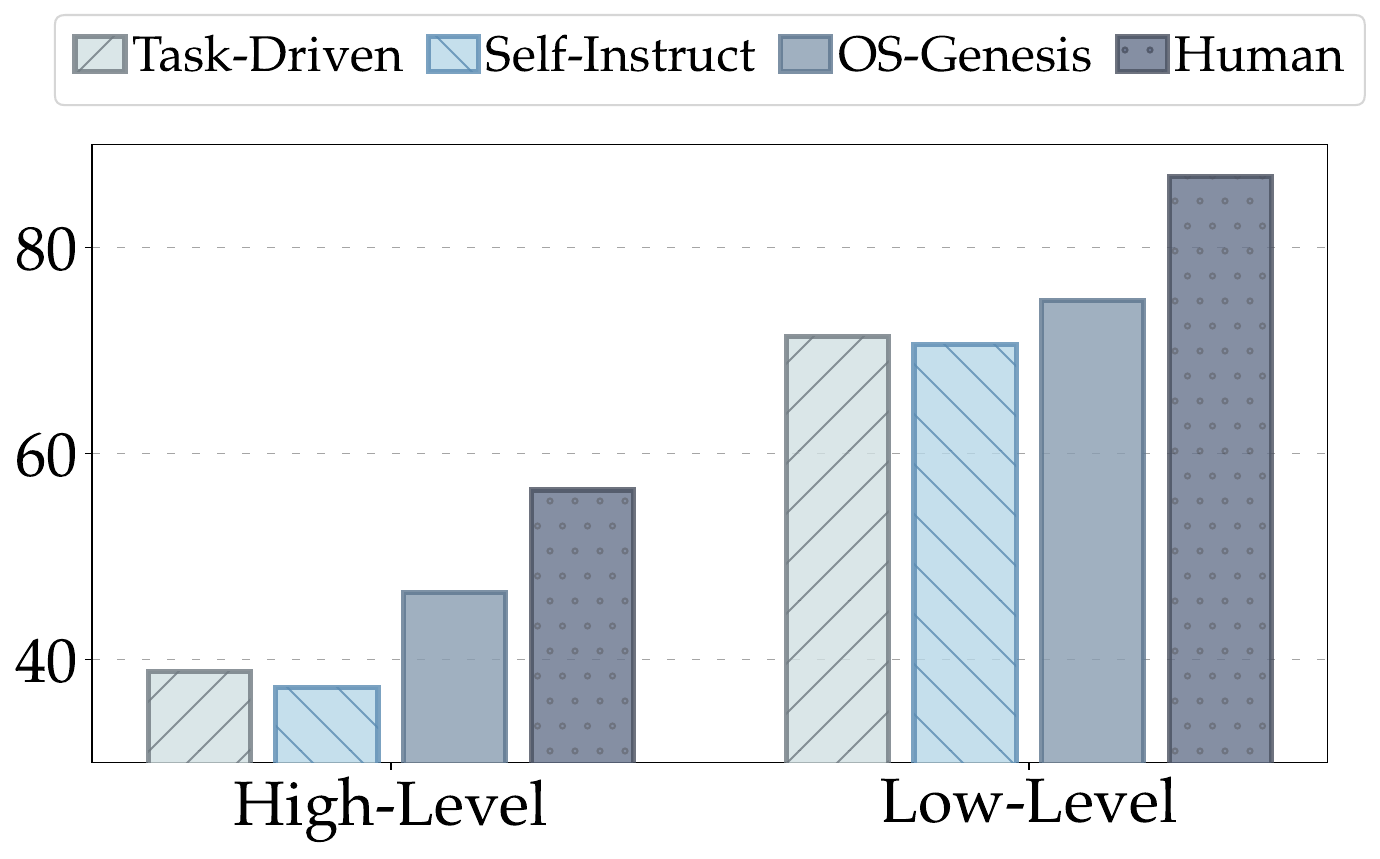}
        \subcaption{Qwen2-VL-7B-Instruct}
    \end{subfigure}
     \vspace{-0.5em}
    \caption{Comparison of training effectiveness between \ours trajectories and human-annotated trajectories.}
    \label{fig:data_diff_sources_analysis}
    \vspace{-0.75em}
\end{figure}
As shown in Figure~\ref{fig:data_diff_sources_analysis}, \ours\ significantly narrows the performance gap between synthetic trajectories and human-annotated trajectories.
This is notably evident in high-level tasks, demonstrating that agents trained on \ours\ trajectories can plan and solve problems more closely aligned with human manners.
In terms of average success rate,
viewing human-annotated data as the gold standard,
the performance retention rate of \ours data surpasses 80\%.
\section{Conclusion}
\label{sec:conc}

We introduce \ours, a data synthesis pipeline to fuel diversified computer control agents.
By leveraging a novel interaction-driven approach, 
\ours overcomes the critical bottlenecks of constructing meaningful and diverse GUI tasks in previous practices.
Through extensive evaluations on challenging online benchmarks,
we demonstrate that \ours-synthesized data has led to a breakthrough in GUI agents’ planning and action capabilities.
Moreover, 
our synthesized trajectories exhibit greater diversity and substantially narrow the quality gap between synthetic data and human annotations.
\ours provides a promising direction for generating high-quality trajectory data for GUI agent training,
bringing the community one step closer to achieving digital automation.

\section*{Limitations}

While \ours demonstrates the potential to overcome critical challenges in acquiring GUI trajectory data,
it is important to acknowledge certain limitations:

\paragraph{Proprietary Models.}
We build our GUI agents upon open-source VLMs, 
but for data quality, we leverage \gpt\ for exploration and reward modeling in the annotation process.
The reason we did not replace this process with open-source counterparts is that existing open-source VLMs lack the ability to follow user instructions and proactively complete exploration in online environments.
We believe that in the future, 
more capable action models can bridge this gap and replace proprietary components in this pipeline.

\paragraph{Data usage.}
Throughout this work, 
we employ textual and visual representations to train and evaluate our GUI agents.
This is designed to (1) maximize agents' planning and action capabilities in semantically rich environments,
and 
(2) ensure evaluation consistency across different environments.
We are aware that using either textual or visual data alone could also contribute to constructing GUI agents,
provided that the I/O format and training strategies are appropriately adjusted.
We leave the partial use of full trajectory data as future works.

\paragraph{Model-based Trajectory Construction.}
We employ a model-based approach to build trajectories, 
thereby eliminating the need for human annotation. 
However, this also results in the quantity of successfully constructed trajectories being somewhat constrained by the capabilities of the task-executing model.
We expect more advanced VLMs with GUI capability to address this in the future.

\section*{Broader Impacts}

Computer agents operating in an OS environment could potentially affect the normal functioning of the system.
However, considering that all settings in this work are conducted within virtual environments, we do not view this as a concern.

\section*{Acknowledgement}
This research is supported by WYNG Foundation (AR25AG100407) and Shanghai Artificial Intelligence Laboratory. 
We thank the reviewers of the SCI-FM Workshop @ ICLR 2025 and ACL Rolling Review for their valuable feedback, which has helped us improve this work.
\section*{Author Contributions}
The authors contributed to this work in the following ways:
\paragraph{Project Leadership.} Qiushi Sun, Kanzhi Cheng, Zhiyong Wu.

\paragraph{OS-Genesis Concept.} Zhiyong Wu.

\paragraph{Data Curation.} Kanzhi Cheng, Qiushi Sun, Fangzhi Xu, Yian Wang, Zichen Ding, Liheng Chen, Chengyou Jia, Zhoumianze Liu, Chuanyang Jin.

\paragraph{Model and Training.} Zichen Ding, Kanzhi Cheng, Qiushi Sun, Zhenyu Wu.

\paragraph{Experiments and Analysis.} Qiushi Sun, Zichen Ding, Yian Wang, Kanzhi Cheng, Chuanyang Jin, Zhenyu Wu, Junxian He.

\paragraph{Paper Writing.} Qiushi Sun, Kanzhi Cheng, Zhiyong Wu, Junxian He, Chuanyang Jin, Zichen Ding.

\paragraph{Demos and Websites.} Qiushi Sun, Zhoumianze Liu.

\paragraph{Discussions.} All authors participate in research discussions and provide insightful technical advice.

\paragraph{Strategic Advice.} Junxian He, Guohao Li, Ben Kao, Yu Qiao.

\bibliography{references}

\appendix

\newpage

\section{Details of Benchmarks}
\label{app:bench}

Here we present more information about the benchmarks involved in evaluating \ours.

\paragraph{AndroidControl.}
AndroidControl~\citep{li2024androidcontrol} is a benchmark designed to evaluate real-world mobile control agents, created from human-collected tasks within the Android environment, consisting of 7,708 tasks across 1,412 trajectories.
It includes two SeqIO tasks: (i) SeqIO HL (high-level), where the prompt contains only a high-level instruction, and (ii) SeqIO LL (low-level), where both a low-level instruction and its corresponding high-level instruction are included.
In terms of evaluation metrics, AndroidControl calculates the success rate (SR) and action type accuracy (Type) based on ground truth action labels.
In our experimental setup, we add the screenshot's accessibility tree and historical actions from the current trajectory as additional observation space to better simulate the agent's execution environment.
In addition, following \citealp{lu2024guiodyssey}, we consider the
coordinates correct if they fall within a distance of 14\% screen width from the ground truth.

\paragraph{AndroidWorld.}
AndroidWorld~\citep{rawles2024androidworld} is an online benchmark for evaluating autonomous agents in Android environments, featuring 116 tasks across 20 real-world apps. Tasks are parameterized with randomized inputs, enabling diverse scenarios and robust evaluations. Success rates (SR) are assessed using system state inspections without modifying the app source code.
Due to app unavailability, a total of 112 tasks are actually used.
Tasks marked as ``NaN'' are re-tested, and those that remain incomplete after re-testing are uniformly marked as false to ensure fair comparisons.

\paragraph{WebArena.}
WebArena~\citep{zhou2024webarena} is a realistic web benchmark for autonomous digital agents, comprising 812 challenging web navigation tasks derived from 241 task templates,
including maps, e-commerce, Reddit forums, and software development.
It features robust evaluation programs that assess the success rate (SR) based on functional correctness.
We follow the standard practices of WebArena by using the default action space (including actions such as clicks and inputs) and employing screenshots and the accessibility tree as the observation space for multimodal GUI agents.
For hosting the online evaluation environment, we use an Amazon EC2 instance (t3a.xlarge, 1000GB EBS root volume). Due to the high cost of evaluation, each task template is tested once, resulting in 241 tests conducted.
\section{Experimental Details}
\label{app:exp_details}

\paragraph{Action Spaces.}
All actions included in the data synthesized by \ours are covered within the types listed in Table~\ref{tab:action_space_mobile} (mobile) and Table~\ref{tab:action_space_web} (web). For AndroidWorld, additional two actions: \texttt{terminate} and \texttt{keyboard\_enter} are incorporated to meet the requirements of evaluation.

\begin{table}[h]
\centering
\resizebox{0.5\textwidth}{!}{%
\begin{tabular}{@{}ll@{}}
\toprule
Action & Description \\
\midrule
\texttt{click} & Clicks at the target elements. \\
\texttt{long\_press} & Presses and holds on the target element. \\
\texttt{type} & Types the specified text at the current cursor location. \\
\texttt{scroll} & Scrolls in a specified direction on the screen. \\
\texttt{navigate\_home} & Navigates to the device’s home screen. \\
\texttt{navigate\_back} & Returns to the previous screen or page. \\
\texttt{open\_app} & Launches the specified application. \\
\texttt{wait} & Agent decides it should wait.  \\
\midrule
\texttt{terminate} & Agent decides the task is finished. \\
\texttt{keyboard\_enter} & Presses the Enter key. \\
\bottomrule
\end{tabular}
}
\caption{Action space for mobile tasks.}  
\label{tab:action_space_mobile}
\end{table}

\begin{table}[h]
\centering
\resizebox{0.5\textwidth}{!}{%
\begin{tabular}{@{}ll@{}}
\toprule
Action & Description \\
\midrule
\texttt{click [id]} & Clicks on an element with a specific id. \\
\texttt{type [id] [content]} & Types the content into the field with id. \\
\texttt{hover [id]} & Hovers on an element with id. \\
\texttt{press [key\_comb]} & Presses the key combination using the keyboard. \\
\texttt{scroll [down|up]} & Scrolls up and down. \\
\texttt{new\_tab} & Opens a new tab. \\
\texttt{tab\_focus [tab\_index]} & Switches the current focus to a specific tab. \\
\texttt{close\_tab} & Closes the current tab. \\
\texttt{goto [url]} & Navigates to a specific URL. \\
\texttt{go\_back} & Navigates to the previous page. \\
\texttt{go\_forward} & Navigates to the next page. \\
\bottomrule
\end{tabular}
}
\caption{Action space for web tasks.}  
\label{tab:action_space_web}
\end{table}

\paragraph{Prompts.}
The instructions we employed for evaluating baselines and \ours on AndroidWorld and AndroidControl are listed in Prompt~\ref{fig:aw-eval-prompt} and Prompt~\ref{fig:ac-eval-prompt} respectively.
\section{Reverse Task Synthesis Details}
\label{app:rev}
Our reverse task synthesis process simulates how humans explore new tasks in an unknown GUI environment. 
After performing actions on random elements, humans infer possible subsequent actions by observing changes on the screen, thus continuing their exploration to construct a complete trajectory for executing a particular task. 
In our reverse task synthesis, we provide \gpt with the current action being executed, before-and-after screenshots of the screen changes, and a red bounding box highlighting the interacted element in the screenshots. 
This allows \gpt to first comprehend the action being performed and then associate the possible high-level task based on the observed screen changes.
The detailed association prompts for synthesizing high-level instruction data for both Android and Web are provided in Prompt~\ref{fig:association-android-prompt} and Prompt~\ref{fig:association-web-prompt} respectively.

\section{Model and Training Details}
\label{app:training_details}

\paragraph{InternVL2-\{4B,8B\}.} InternVL2~\citep{chen2023internvl} utilizes Dynamic Aspect Ratio Matching to handle dynamic high-resolution inputs. 
In our training setting, we set the \texttt{max\_dynamic\_patch} parameter to 24 to comprehensively capture the fine-grained details of the image.
Consequently, the resized input image is partitioned into a maximum of 24 tiles, each of 448×448 pixels, while a thumbnail of the entire image is included to preserve global contextual information.
\paragraph{Qwen2-VL-7B-Instruct.}
Qwen2-VL~\citep{Qwen2VL} introduces the Naive Dynamic Resolution mechanism, which is capable of handling images of any resolution by mapping them into a dynamic number of visual tokens, providing a more human-like visual processing experience.
Through our experiments, we found that configuring the \texttt{image\_resolution} parameter to 1024 for both training and inference produces outstanding results in GUI agent tasks, while also contributing to the optimization of the model's training and inference costs.
\paragraph{Accessibility Tree.}
The accessibility tree represents the hierarchical relationships and attributes of all interactive or accessible elements on a screen, providing rich GUI information in text form to train GUI agents.
In constructing the training data, we filter the accessibility tree to retain only the position or index information of elements visible on the screen, reducing the interference of excessive redundant text in model training.
\paragraph{Data Format.}
We follow the data formats of AndroidWorld and WebArena to construct our training data, ensuring consistency in formatting between the training and evaluation phases.
The detailed training instructions for Android and Web data are listed in Prompt~\ref{fig:ac-train-prompt} and Prompt~\ref{fig:web-train-prompt} respectively.

\begin{figure*}[ht]
    \centering
    \begin{subfigure}[b]{0.31\textwidth}
        \centering
        \includegraphics[width=\textwidth]{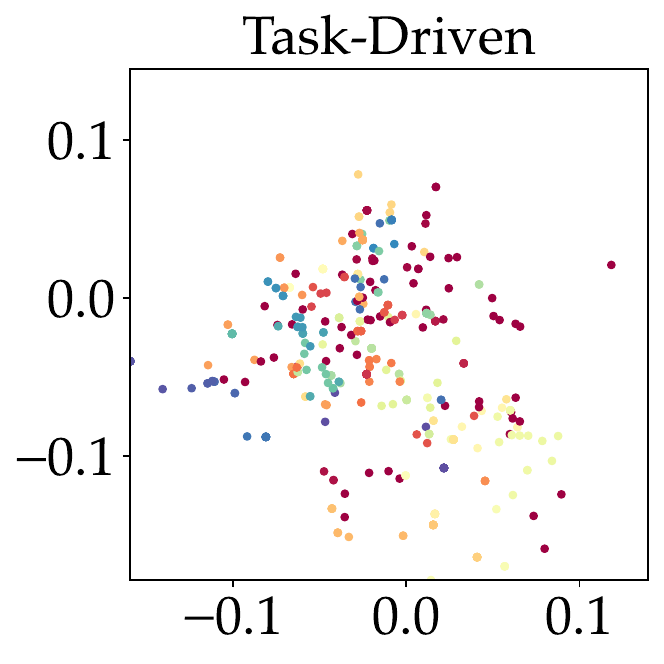}
        \label{fig:data_diversity_visualization1}
    \end{subfigure}
    \hfill
    \begin{subfigure}[b]{0.31\textwidth}
        \centering
        \includegraphics[width=\textwidth]{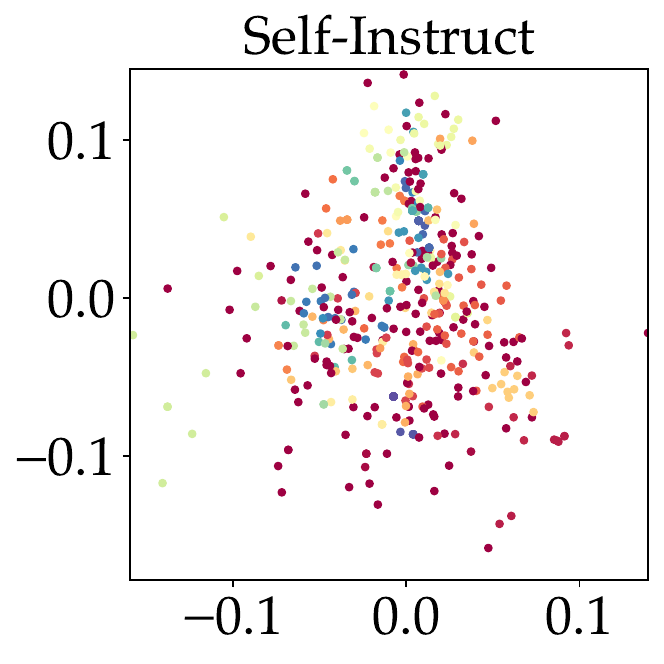}
        \label{fig:data_diversity_visualization2}
    \end{subfigure}
    \hfill
    \begin{subfigure}[b]{0.31\textwidth}
        \centering
        \includegraphics[width=\textwidth]{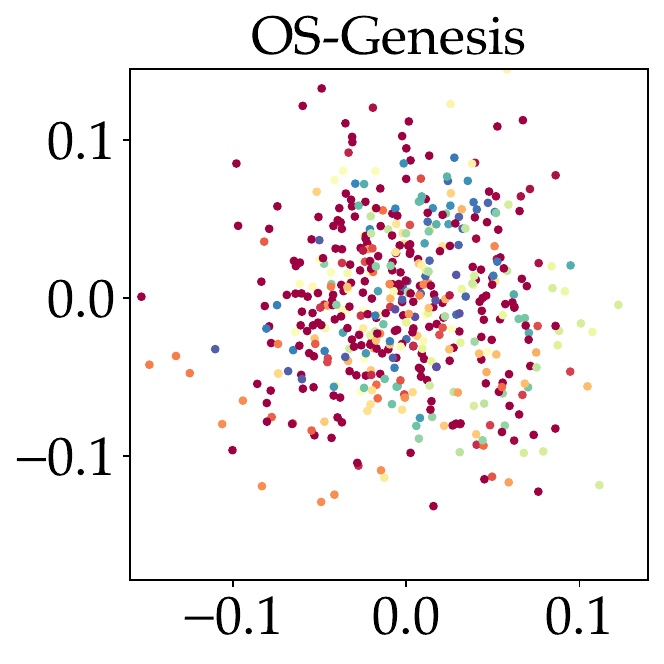}
        \label{fig:data_diversity_visualization_3}
    \end{subfigure}
    \vspace{-0.5em}
    \caption{Visualization of the instruction embeddings across various synthetic datasets.}
    \label{fig:data_diversity_visualization}
\end{figure*}

\section{Baseline Settings}
\label{sec:baseline}

\subsection{Task-Driven}
\label{app:bsae:task}
\begin{figure*}[htbp]
    \centering
    \begin{subfigure}{0.3\textwidth}
        \centering
        \includegraphics[width=0.9\linewidth]{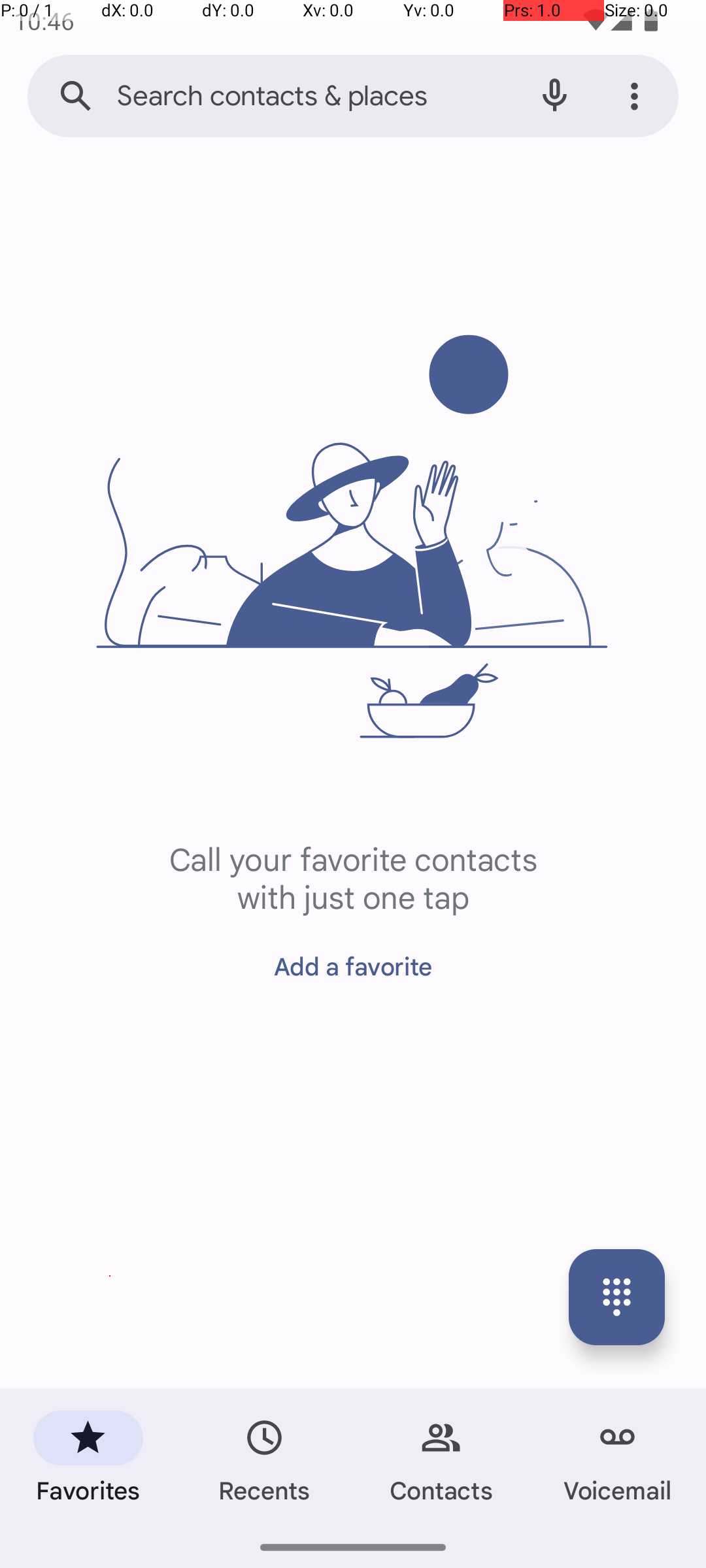}
        \caption{Contacts}
        \label{fig:mobile_subfig1}
    \end{subfigure}
    \hfill 
    \begin{subfigure}{0.3\textwidth}
        \centering
        \includegraphics[width=0.9\linewidth]{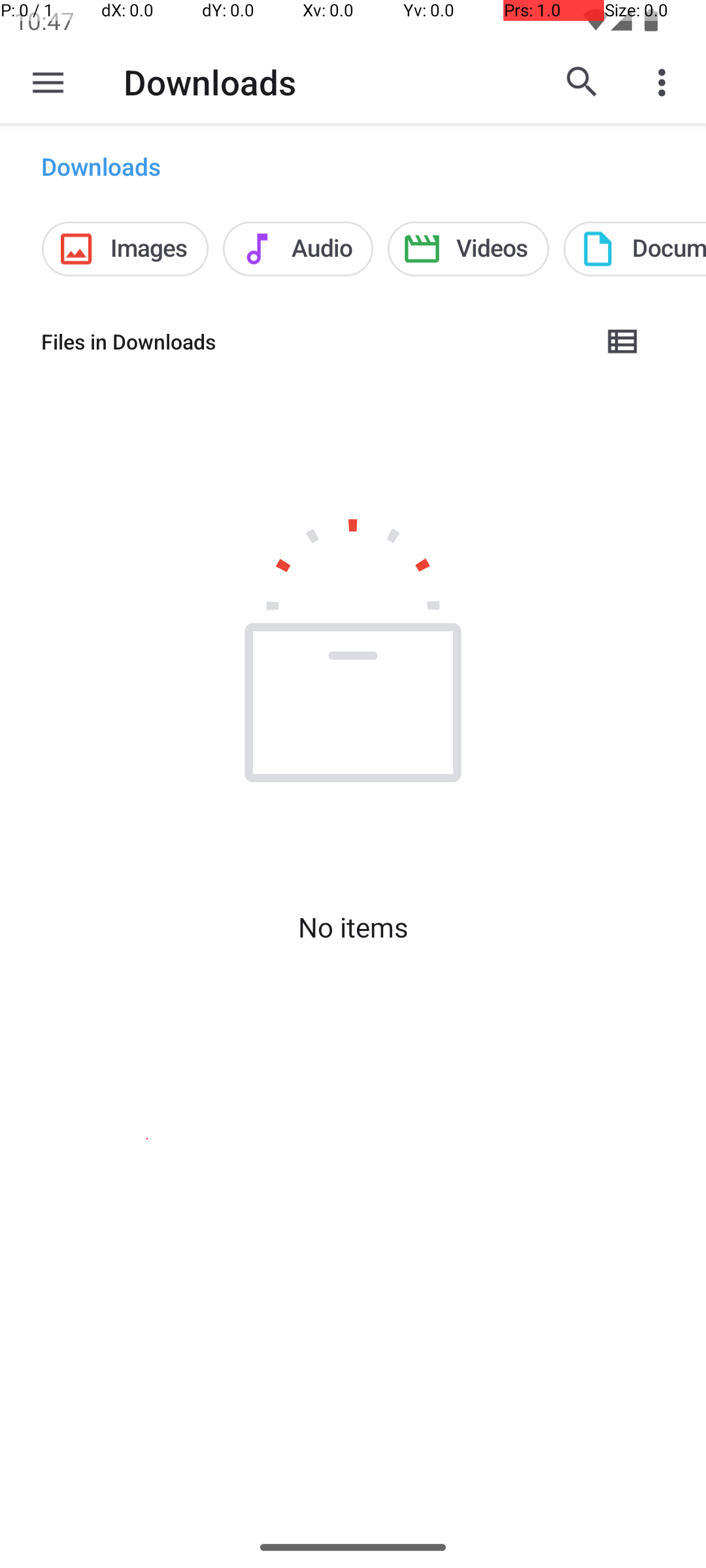}
        \caption{Files}
        \label{fig:mobile_subfig2}
    \end{subfigure}
    \hfill
    \begin{subfigure}{0.3\textwidth}
        \centering
        \includegraphics[width=0.9\linewidth]{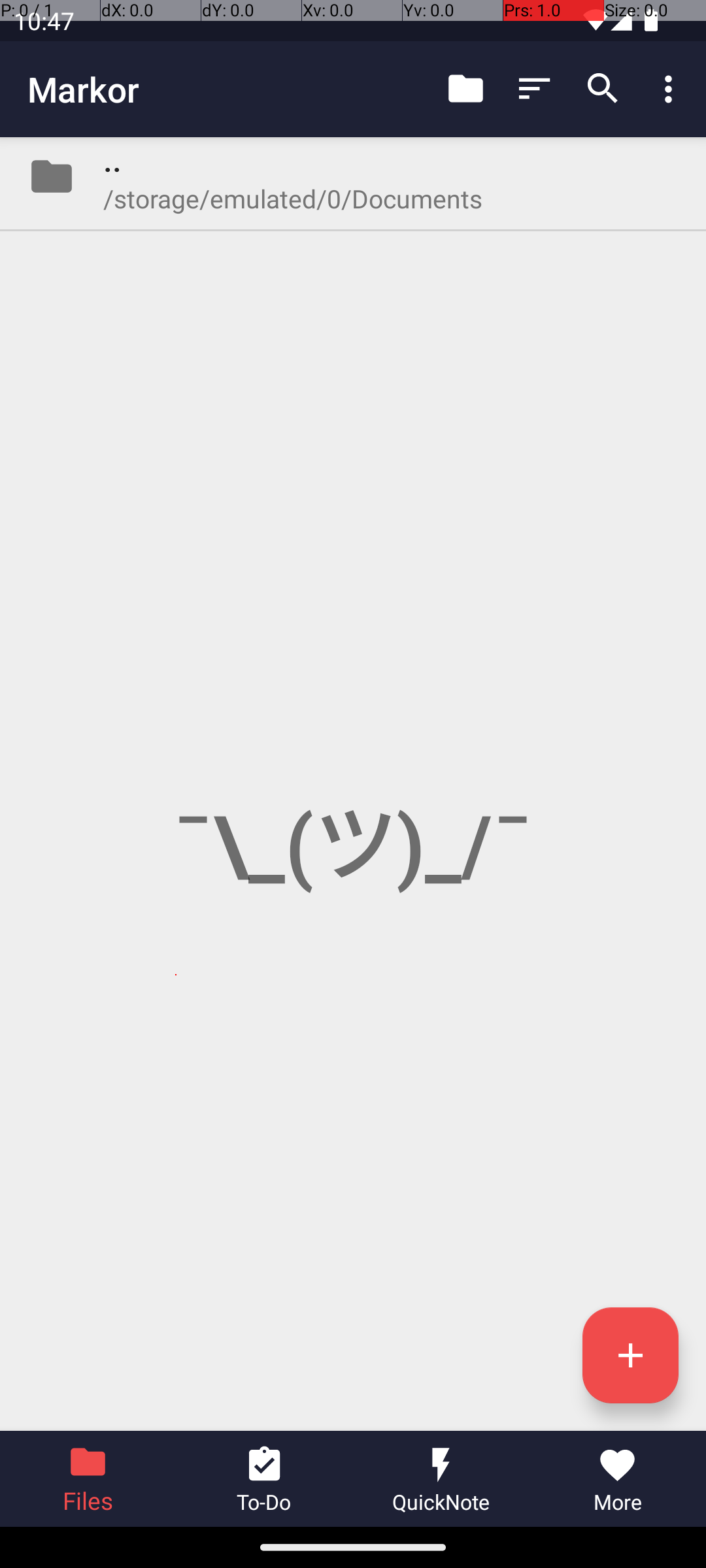}
        \caption{Markor}
        \label{fig:mobile_subfig3}
    \end{subfigure}
    \caption{Examples of initial screens employed in building task-driven baselines for mobile tasks.}
    \label{fig:baseline_mobile_init}
\end{figure*}

\begin{figure*}[htbp]
    \centering
    \begin{subfigure}{0.31\textwidth}
        \centering
        \includegraphics[width=\linewidth]{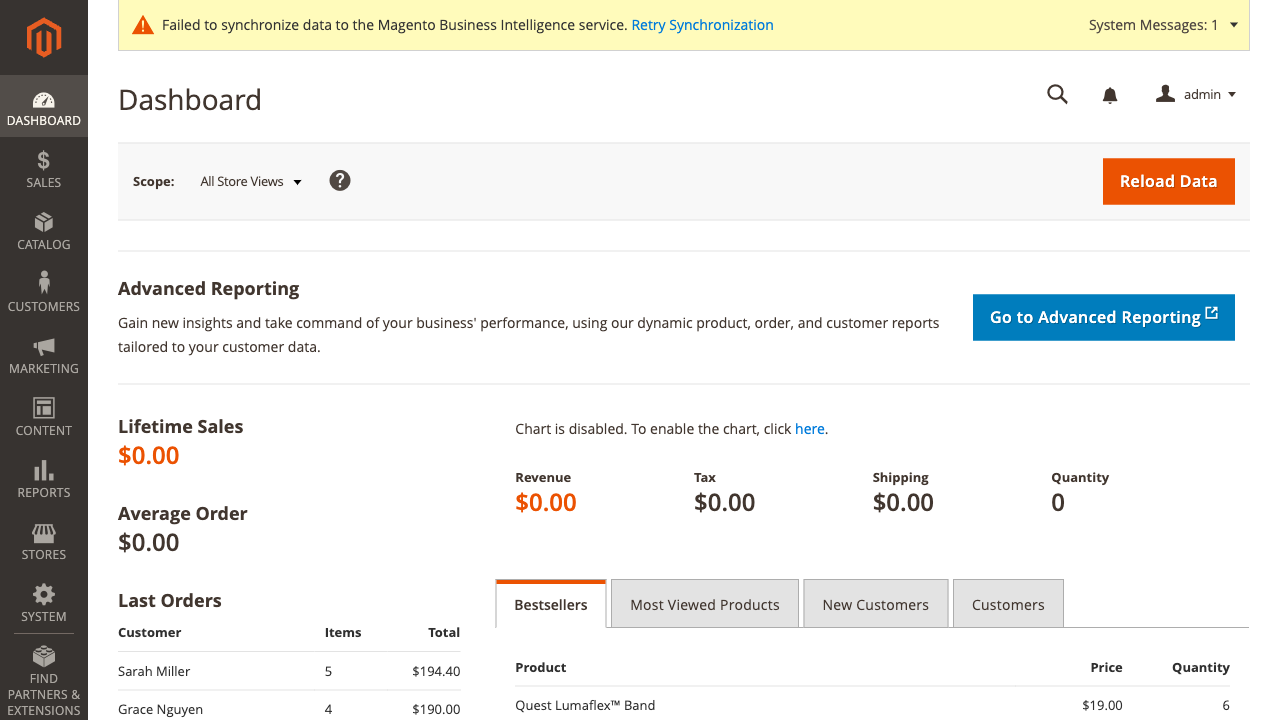}
        \caption{CMS}
        \label{fig:web_subfig1}
    \end{subfigure}
    \hfill 
    \begin{subfigure}{0.31\textwidth}
        \centering
        \includegraphics[width=\linewidth]{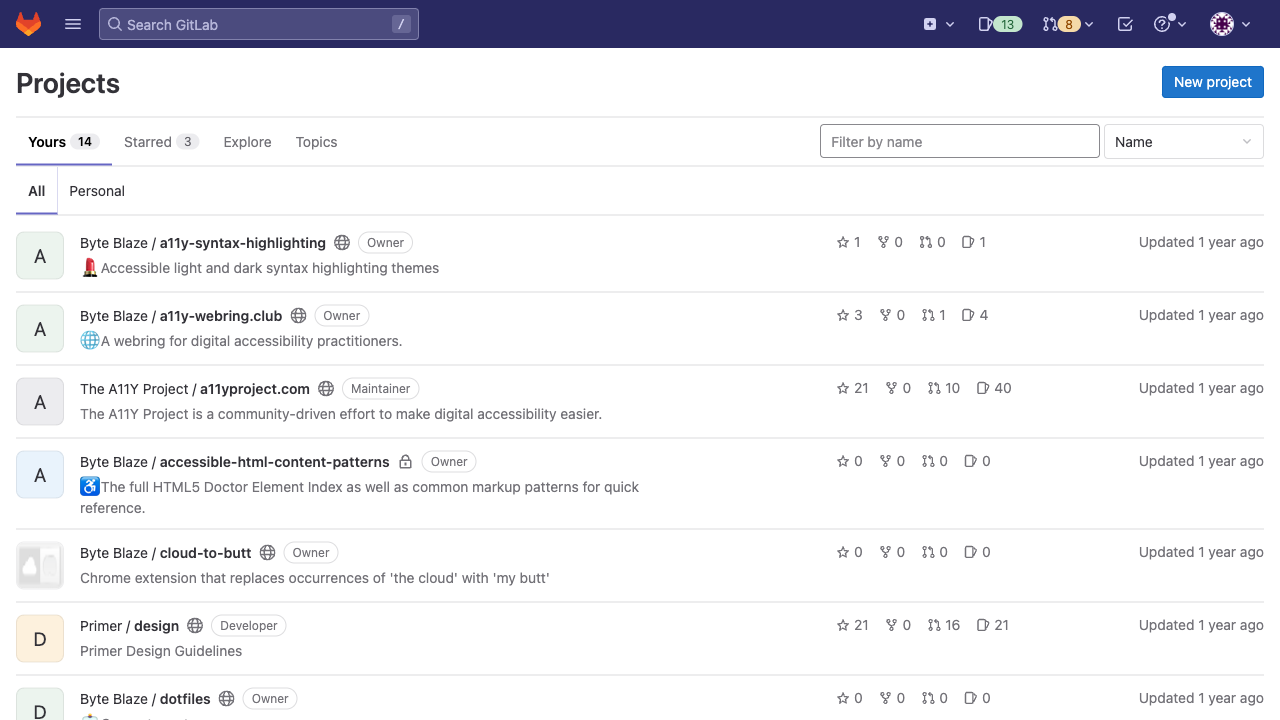}
        \caption{GitLab}
        \label{fig:web_subfig2}
    \end{subfigure}
    \hfill
    \begin{subfigure}{0.31\textwidth}
        \centering
        \includegraphics[width=\linewidth]{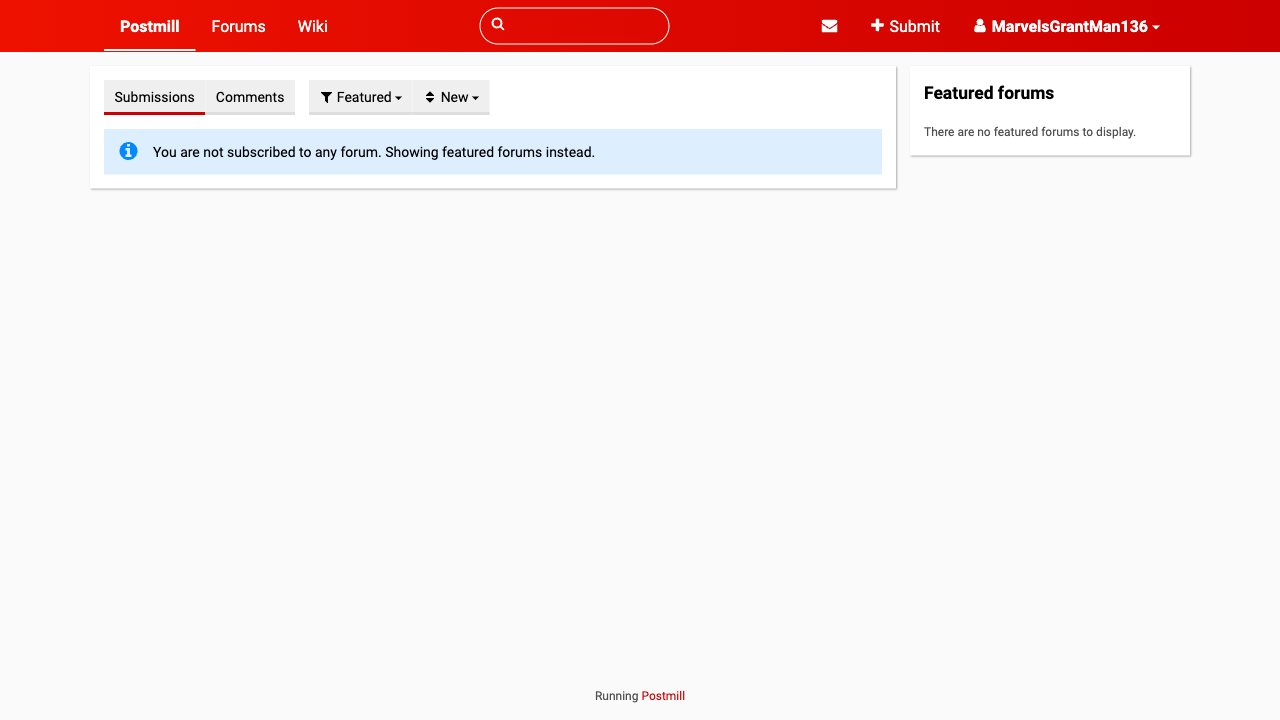}
        \caption{Reddit}
        \label{fig:web_subfig3}
    \end{subfigure}
    \caption{Examples of initial screens employed in building task-driven baselines for web tasks.}
    \label{fig:baseline_web_init}
\end{figure*}

Following prior work \cite{he2024openwebvoyager, lai2024autowebglm} on collecting tasks for GUI agents, we guide \gpt to infer possible high-level instructions based on the initial GUI interface (\textit{e.g.}, the homepage of a social forum like Reddit).
Some examples of initial screens are demonstrated in Figure~\ref{fig:baseline_mobile_init} (mobile) and Figure~\ref{fig:baseline_web_init} (web).

\subsection{Task-Driven w. Self Instruct}

Building upon the task-driven baseline in \ref{app:bsae:task},
we incorporate self-instruction~\citep{wang2023selfinstruct} data as a second baseline.
This is constructed by randomly sampling 3 demonstrations from the above task-driven high-level instructions as in-context examples for each synthesis iteration.

Notably, we make certain that the total number of trajectories for the baseline is at least equal to that of our method to avoid data imbalance and maintain fairness in comparisons.

\section{Details of Trajectory Reward Model}
\label{app:trm}

The Trajectory Reward Model (TRM) primarily assesses the quality of agent trajectories by focusing on completion and coherence.
Based on a high-level instruction to complete, the agent's entire action history (\textit{e.g.}, low-level instructions), and screenshots from the last three timesteps, \gpt is prompted to assign a score between 1 and 5 for the trajectory.
Instead of instruction and in-context learning~\citep{sun2024bbt}, we include in the prompt specific aspects of coherence and completion to consider, along with detailed descriptions of what each score from 1 to 5 represents.
Given the similarity between mobile and web tasks, we apply the same TRM to both, as shown in prompt ~\ref{fig:trm-prompt-mobile}.
Recognizing the critical role that TRM plays in ensuring data quality, additional analyses are conducted to validate its robustness.

\subsection{Model Scoring Alignment}
To demonstrate that models with varying capabilities can effectively serve as the TRM, we adopt Qwen2.5-VL-72B~\citep{Qwen25VL} as the TRM backbone and perform additional experiments on a small-scale dataset.
Specifically, 
we sample 200 web and mobile trajectories for evaluation,
results are shown in Table~\ref{tab:trm_qwen}.

\begin{table}[h]
\centering
\begin{tabular}{lcc}
\toprule
\textbf{Domain} & \textbf{Spearman Corr.} & \textbf{p-val} \\
\midrule
Web    & 0.788 & 2.16e-22 \\
Mobile & 0.729 & 7.82e-18 \\
\bottomrule
\end{tabular}
\caption{Consistency between \gpt and Qwen2.5-VL-72B as the backbone of TRM.}
\label{tab:trm_qwen}
\end{table}

The results demonstrate that \ours can also work effectively with open-source models.

\subsection{Human Scoring Alignment}
To further validate the TRM, we sample 100 web and mobile trajectories and invite human annotators to label them according to the same guidelines used for the GPT-4o-based TRM. Subsequently, we calculated the Spearman correlation coefficients between the human annotations and the TRM-assigned rewards, as presented below.

\begin{table}[h]
\centering
\begin{tabular}{lcc}
\toprule
\textbf{Domain} & \textbf{Spearman Corr.} & \textbf{p-val} \\
\midrule
Mobile & 0.813 & 9.23e-24 \\
Web    & 0.798 & 2.54e-23 \\
\bottomrule
\end{tabular}
\caption{Consistency between TRM and human annotators for reward modeling.}
\label{tab:trm_human}
\end{table}

The results also demonstrate that TRM aligns well with human evaluations.



\begin{figure*}
    \centering
    \setlength{\fboxrule}{0.85pt}
    \fbox{ \footnotesize
        \parbox{1\textwidth}{\texttt{\textbf{Prompt for Planning Training}\\
\texttt{<image>} \\
You are a GUI task expert, I will provide you with a high-level instruction, an action history, \\
a screenshot with its corresponding accessibility tree. \\
\textbf{High-level instruction:} \{high\_level\_instruction\} \\
\textbf{Action history:} \{action\_history\} \\
\textbf{Accessibility tree:} \{a11y\_tree\} \\
Please generate the low-level thought and action for the next step. 
        }
    }}
    \fbox{ \footnotesize 
        \parbox{1.0\textwidth}{\texttt{\textbf{Prompt for Action Training}\\
\texttt{<image>} \\
You are a GUI task expert, I will provide you with an action history, 
a screenshot with its corresponding accessibility tree, and a low-level thought. \\
\textbf{Action history:} \{action\_history\} \\
\textbf{Accessibility tree:} \{a11y\_tree\} \\
\textbf{Low-level thought:} \{low\_level\_thought\} \\
Please generate the action for the next step. 
}}}
    \captionsetup{labelformat=default, name=Prompt}
    \caption{Prompts for training our agents on Android.}
    \label{fig:ac-train-prompt}
\end{figure*}
\begin{figure*}
    \centering
    \setlength{\fboxrule}{0.85pt}
    \fbox{ \footnotesize
        \parbox{1\textwidth}{\texttt{\textbf{Prompt for Planning Training}\\
\texttt{<image>} \\
**Task Description** \\
You are an intelligent agent completing web-based tasks. \\
Based on the user's objective (i.e. instruction), current interface information (i.e. screenshot and its corresponding accessibility tree), and action history, determine the next action. \\
\\
**Available Actions** \\
- \texttt{click [id]}: This action clicks on an element with a specific id on the webpage. \\ 
- \texttt{type [id] [content] [press\_enter\_after=0|1]}: Use this to type the content into the field with id. By default, the "Enter" key is pressed after typing unless \texttt{press\_enter\_after} is set to 0.  \\
- \texttt{hover [id]}: Hover over an element with id.  \\
- \texttt{press [key\_comb]}: Simulates the pressing of a key combination on the keyboard (e.g., Ctrl+v).  \\
- \texttt{scroll [direction=down|up]}: Scroll the page up or down.  \\
- \texttt{new\_tab}: Open a new, empty browser tab.  \\
- \texttt{tab\_focus [tab\_index]}: Switch the browser's focus to a specific tab using its index.  \\
- \texttt{close\_tab}: Close the currently active tab.  \\
- \texttt{goto [url]}: Navigate to a specific URL.  \\
- \texttt{go\_back}: Navigate to the previously viewed page.  \\
- \texttt{go\_forward}: Navigate to the next page (if a previous \texttt{go\_back} action was performed).  \\
- \texttt{stop [answer]}: Issue this action when you believe the task is complete. If the objective is to find a text-based answer, provide the answer in the bracket. If you believe the task is impossible to complete, provide the answer as ``N/A'' in the bracket.\\
\\
**Output Format** \\
First, generate the reasoning process for the action. Then, generate the action in the correct format. Start with a ``In summary, the next action I will perform is'' phrase, followed by action inside \texttt{\`}\texttt{\`}\texttt{\`}\texttt{\`}\texttt{\`}\texttt{\`}. \\
For example: \\
``Let's think step-by-step. To add a product to the shopping cart, I need to navigate to the catalog or product section. 
The ``CATALOG'' link is available with ID [1234]. 
In summary, the next action I will perform is \texttt{\`}\texttt{\`}\texttt{\`}click [1234]\texttt{\`}\texttt{\`}\texttt{\`}''. \\
\\
\textbf{Instruction:} \{instruction\} \\
\textbf{Accessibility tree:} \{a11y\_tree\} \\
\textbf{Action History:} \{action\_history\} \\
What's the next action?
    }}}
    \fbox{ \footnotesize 
        \parbox{1.0\textwidth}{\texttt{\textbf{Prompt for Action Training}\\
\texttt{<image>} \\
You are an intelligent agent completing web-based tasks. I will provide you with available actions, a screenshot with its corresponding accessibility tree, and a low-level thought. \\
\\
**Available Actions** \\
- \texttt{click [id]}: This action clicks on an element with a specific id on the webpage. \\ 
- \texttt{type [id] [content] [press\_enter\_after=0|1]}: Use this to type the content into the field with id. By default, the ``Enter'' key is pressed after typing unless \texttt{press\_enter\_after} is set to 0.  \\
- \texttt{hover [id]}: Hover over an element with id.  \\
- \texttt{press [key\_comb]}: Simulates the pressing of a key combination on the keyboard (e.g., Ctrl+v).  \\
- \texttt{scroll [direction=down|up]}: Scroll the page up or down.  \\
- \texttt{new\_tab}: Open a new, empty browser tab.  \\
- \texttt{tab\_focus [tab\_index]}: Switch the browser's focus to a specific tab using its index.  \\
- \texttt{close\_tab}: Close the currently active tab.  \\
- \texttt{goto [url]}: Navigate to a specific URL.  \\
- \texttt{go\_back}: Navigate to the previously viewed page.  \\
- \texttt{go\_forward}: Navigate to the next page (if a previous \texttt{go\_back} action was performed).  \\
- \texttt{stop [answer]}: Issue this action when you believe the task is complete. If the objective is to find a text-based answer, provide the answer in the bracket. If you believe the task is impossible to complete, provide the answer as ``N/A'' in the bracket.\\
\\
\textbf{Accessibility tree:} \{a11y\_tree\} \\
\textbf{Low-level thought:} \{low\_level\_thought\} \\
Please generate the action inside \texttt{\`}\texttt{\`}\texttt{\`}\texttt{\`}\texttt{\`}\texttt{\`} for the next step. 
}}}
    \captionsetup{labelformat=default, name=Prompt}
    \caption{Prompts for training our agents on Web.}
    \label{fig:web-train-prompt}
\end{figure*}

\begin{figure*}
    \centering
    \setlength{\fboxrule}{0.85pt}
    \fbox{ \footnotesize
        \parbox{1\textwidth}{\texttt{\textbf{Prompt for Associating High-Level Tasks}\\
        \\
You are an expert at envisioning specific tasks corresponding to changes in mobile screenshots. I will provide you with the following:\\
1. The type of action currently being executed. The type of action currently being executed, which can be one of five types: \texttt{CLICK}, \texttt{SCROLL}, \texttt{TYPE}, \texttt{PRESS\_BACK}, and \texttt{LONG\_PRESS}. If the action is \texttt{TYPE}, an additional value representing the input will be provided. If the action is \texttt{SCROLL}, an additional scroll direction will be provided.\\
2. Screenshots of the interface before and after the current action is performed. If the action is \texttt{CLICK}, the pre-action screenshot will include a red bbox highlighting the element being interacted with (if applicable). Pay particular attention to the content of the element corresponding to the red bbox.\\
3. The name of the app where the current screenshot is located.\\
\\
Your task is to envision a specific task based on the current action and the corresponding changes in screenshots. The output should include three parts:\\
\textbf{1. Sub-Instruction:} Based on the interface change caused by the current action, generate a corresponding natural language instruction for the current action. The instruction should be concise, clear, and executable. It must include specific details critical to the operation, such as file names, times, or other content as they appear in the screenshots. For example: ``Scroll left to open the app drawer, displaying all installed applications on the devic'', ``Click the chat interface, allowing the user to view and participate in conversation'', ``Type the username `Agent', preparing for the next step in logging into the account''.\\
\textbf{2. Analysis:} Based on the interface changes and the current action instructions, analyze the possible subsequent operations. This analysis should involve step-by-step reasoning, considering the potential changes on the screen and the actions that can be taken after these changes. For example: ``After clicking the plus button, a dropdown menu appears with an option to create a document. I can select this option to create a new document. First, I need to name the document, then enter any content into the document, and finally save the document and exit''.\\
\textbf{3. High-Level-Instruction:} Based on the analysis results, envision a high-level task that can be completed within the current interface. There are two types of High-Level-Instruction:\\
Task-Oriented: Completing a series of operations to achieve a specific goal.\\
Question-Oriented: Performing a series of operations and deriving an answer to a specific question.\\
For example: \{examples\}.\\
Ensure that the High-Level-Instruction is executable by including all critical specifics, such as file names, relevant timings, or required details.\\
\\
You ONLY need to return a dictionary formatted as follows:\\
\{\\
  ``Sub-Instruction'': ``xxx'', \\
  ``Analysis'': ``xxx'', \\
  ``High-Level-Instruction'': ``xxx''\\
\}\\
\\
\textbf{Current Action:} \{current\_action\} \\
\textbf{App Name:} \{app\_name\} \\
RETURN ME THE DICTIONARY I ASKED FOR.
}
    }}
    \captionsetup{labelformat=default, name=Prompt} 
    \caption{Prompts for associating high-level tasks on mobile.}
    \label{fig:association-android-prompt}
\end{figure*}

\begin{figure*}
    \centering
    \setlength{\fboxrule}{0.85pt}
    \fbox{ \footnotesize
        \parbox{1\textwidth}{\texttt{\textbf{Prompt for Associating High-Level Tasks}\\
        \\
You are a GUI (Graphical User Interface) expert capable of analyzing interface changes and envisioning executable tasks or instructions. Given a GUI interface change caused by an action (e.g., clicking or typing) and the corresponding element highlighted in red boxes, you are required to analyze the interface and generate related tasks.\\
Your task is to envision tasks based on the current action and the resulting changes in the screenshots. The output should include three components:\\
1. \textbf{Sub-Instruction}: Create a natural language instruction for the current action based on the interface changes it caused. The instruction should be concise, clear, and actionable, incorporating specific details critical to the task, such as elements, file names, timestamps, or other relevant content visible in the screenshots. For example: \\
   - ``Click on the `Add to Cart' button next to the product to add it to your shopping cart.''  \\
   - ``Type `OpenAI' into the search bar to find relevant articles.''  \\
   - ``Scroll down to view the latest blog posts on the homepage.''\\
2. \textbf{Analysis}: Carefully analyze the before-and-after screenshots step by step, focusing on the changes caused by the action. Then, examine key elements in both screenshots and consider possible operations based on these elements. For example: ``The previous screen displayed the main interface of a shopping website, featuring multiple product categories and several showcased items. After clicking the `Sign Up' button, the interface transitioned to a login page where an email and password can be entered to log into an account. The login page also provides other options, such as recovering a password, creating a new account, or logging in with a Google account''.\\
3. \textbf{High-Level Instruction}: Based on the before-and-after screenshots, the action, and the analysis, generate a high-level task that you believe can be completed within the current interface. There are three types of tasks:  \\
   - Information seeking: The user wants to obtain certain information from the webpage, such as product details, reviews, map information, or route comparisons. Please propose clear and specific questions that need an explicit answer, and avoid asking for summary-type questions, such as ``summarize the information about a product''.  \\
   - Site navigation: The user wants to navigate to a specific page or state.  \\
   - Content modification: The user wants to modify the content of a webpage or its settings.\\
The high-level instruction should be creative. You need to deeply analyze the elements and executable actions on the interface to generate realistic, valuable, and executable tasks that can be completed within the current GUI. The instruction should be specific, actionable, and goal-oriented, ensuring the task can be completed on the current GUI by including all critical specifics such as file names, relevant timings, or required details.\\
\\
Below is a brief description of the current website:  
\{website\_intro\}
\\
Here are some examples of High-Level Instruction for reference:  
\{task\_examples\}
\\
Please generate tasks that can be completed on the current platform, and avoid tasks that are unrelated to the current website.\\
\\
You ONLY need to return a dictionary formatted as follows:\\
\{\\
  ``Sub-Instruction'': ``xxx'', \\
  ``Analysis'': ``xxx'', \\
  ``High-Level-Instruction'': ``xxx''\\
\}\\
\\
\textbf{Current Action:} \{current\_action\} \\
\textbf{Website Name:} \{website\_name\} \\
RETURN ME THE DICTIONARY I ASKED FOR.
}
    }}
    \captionsetup{labelformat=default, name=Prompt} 
    \caption{Prompts for associating high-level tasks on web.}
    \label{fig:association-web-prompt}
\end{figure*}
\begin{figure*}
    \centering
    \setlength{\fboxrule}{0.85pt}
    \fbox{ \footnotesize
        \parbox{1\textwidth}{\texttt{\textbf{Evaluation Prompt for AndroidWorld}\\
        \\
You are a GUI task expert, I will provide you with a high-level instruction, an action history, \\
a screenshot with its corresponding accessibility tree. \\
\\
\textbf{High-level instruction:} \{high\_level\_instruction\} \\
\textbf{Action history:} \{action\_history\} \\
\textbf{Accessibility tree:} \{a11y\_tree\} \\
\\
Please generate the low-level thought and action for the next step.
        }
    }}
    \captionsetup{labelformat=default, name=Prompt} 
    \caption{Prompts for evaluating our agents on AndroidWorld.}
    \label{fig:aw-eval-prompt}
\end{figure*}

\begin{figure*}
    \centering
    \setlength{\fboxrule}{0.85pt}
    \fbox{ \footnotesize
        \parbox{1\textwidth}{\texttt{\textbf{Evaluation Prompt for AndroidControl: High-Level Settings}\\
\texttt{<image>} \\
You are a GUI task expert, I will provide you with a high-level instruction, an action history, \\
a screenshot with its corresponding accessibility tree. \\
\\
\textbf{High-level instruction:} \{high\_level\_instruction\} \\
\textbf{Action history:} \{action\_history\} \\
\textbf{Accessibility tree:} \{a11y\_tree\} \\
\\
Please generate the low-level thought and action for the next step.
        }
    }}
    \fbox{ \footnotesize 
        \parbox{1.0\textwidth}{\texttt{\textbf{Evaluation Prompt for AndroidControl: Low-Level Settings}\\
\texttt{<image>} \\
You are a GUI task expert, I will provide you with a high-level instruction, an action history, \\
a screenshot with its corresponding accessibility tree, and a low-level thought. \\
\\
\textbf{High-level instruction:} \{high\_level\_instruction\} \\
\textbf{Action history:} \{action\_history\} \\
\textbf{Accessibility tree:} \{a11y\_tree\} \\
\textbf{Low-level thought:} \{low\_level\_thought\} \\
\\
Please generate the action for the next step.
}}}
    \captionsetup{labelformat=default, name=Prompt}
    \caption{Prompts for evaluating our agents on AndroidControl.}
    \label{fig:ac-eval-prompt}
\end{figure*}

\begin{figure*}
    \centering
    \setlength{\fboxrule}{0.85pt}
    \fbox{ \footnotesize
        \parbox{1\textwidth}{\texttt{\textbf{Evaluation Prompt for AndroidControl: High-Level Settings}\\
\texttt{<image>} \\
You are a GUI task expert, I will provide you with a high-level instruction, an action history, \\
a screenshot with its corresponding accessibility tree. \\
\\
\textbf{High-level instruction:} \{high\_level\_instruction\} \\
\textbf{Action history:} \{action\_history\} \\
\textbf{Accessibility tree:} \{a11y\_tree\} \\
\\
Please generate the low-level thought and action for the next step. \\
Candidate Actions: \\
\texttt{``action\_type'': ``type'', ``text'': <text\_input>, ``x'': <x\_coordinate>, ``y'': <y\_coordinate>} \\
\texttt{``action\_type'': ``navigate\_home''} \\
\texttt{``action\_type'': ``navigate\_back''} \\
\texttt{``action\_type'': ``scroll'', ``direction'': <up, down, left, or right>} \\
\texttt{``action\_type'': ``open\_app'', ``app\_name'': <app\_name>} \\
\texttt{``action\_type'': ``wait''} \\
\texttt{``action\_type'': ``dismiss'', ``x'': <x\_coordinate>, ``y'': <y\_coordinate>} \\
\texttt{``action\_type'': ``long\_press'', ``x'': <x\_coordinate>, ``y'': <y\_coordinate>} \\
\texttt{``action\_type'': ``get\_text'', ``x'': <x\_coordinate>, ``y'': <y\_coordinate>} \\
You need to generate a script in the form: \\
\texttt{thoughts:} \\
\texttt{\{THOUGHTS\}} \\
\texttt{actions:} \\
\texttt{\{ACTION\}} \\
Make sure to consider the details in the screenshot and the task requirements to create an accurate and functional script.
        }
    }}
    \fbox{ \footnotesize 
        \parbox{1.0\textwidth}{\texttt{\textbf{Evaluation Prompt for AndroidControl: Low-Level Settings}\\
\texttt{<image>} \\
You are a GUI task expert, I will provide you with a high-level instruction, an action history, \\
a screenshot with its corresponding accessibility tree, and a low-level thought. \\
\\
\textbf{High-level instruction:} \{high\_level\_instruction\} \\
\textbf{Action history:} \{action\_history\} \\
\textbf{Accessibility tree:} \{a11y\_tree\} \\
\textbf{Low-level thought:} \{low\_level\_thought\} \\
\\
Please generate the action for the next step. \\
Candidate Actions: \\
\texttt{``action\_type'': ``type'', ``text'': <text\_input>, ``x'': <x\_coordinate>, ``y'': <y\_coordinate>} \\
\texttt{``action\_type'': ``navigate\_home''} \\
\texttt{``action\_type'': ``navigate\_back''} \\
\texttt{``action\_type'': ``scroll'', ``direction'': <up, down, left, or right>} \\
\texttt{``action\_type'': ``open\_app'', ``app\_name'': <app\_name>} \\
\texttt{``action\_type'': ``wait''} \\
\texttt{``action\_type'': ``dismiss'', ``x'': <x\_coordinate>, ``y'': <y\_coordinate>} \\
\texttt{``action\_type'': ``long\_press'', ``x'': <x\_coordinate>, ``y'': <y\_coordinate>} \\
\texttt{``action\_type'': ``get\_text'', ``x'': <x\_coordinate>, ``y'': <y\_coordinate>} \\
You need to generate a script in the form: \\
\texttt{thoughts:} \\
\texttt{\{THOUGHTS\}} \\
\texttt{actions:} \\
\texttt{\{ACTION\}} \\
Make sure to consider the details in the screenshot and the task requirements to create an accurate and functional script.
}}}
    \captionsetup{labelformat=default, name=Prompt}
    \caption{Prompts for evaluating base models (Zero-Shot) on AndroidControl.}
    \label{fig:ac-eval-prompt-zs}
\end{figure*}
\begin{figure*}
    \centering
    \setlength{\fboxrule}{0.85pt}
    \fbox{ \footnotesize
        \parbox{1\textwidth}{\texttt{\textbf{Evaluation Prompt for WebArena}\\
\texttt{<image>} \\
**Task Description** \\
You are an intelligent agent completing web-based tasks. \\
Based on the user's objective (i.e. instruction), current interface information (i.e. screenshot and its corresponding accessibility tree), and action history, determine the next action. \\
\\
**Available Actions** \\
- \texttt{click [id]}: This action clicks on an element with a specific id on the webpage. \\ 
- \texttt{type [id] [content] [press\_enter\_after=0|1]}: Use this to type the content into the field with id. By default, the ``Enter'' key is pressed after typing unless \texttt{press\_enter\_after} is set to 0.  \\
- \texttt{hover [id]}: Hover over an element with id.  \\
- \texttt{press [key\_comb]}: Simulates the pressing of a key combination on the keyboard (e.g., Ctrl+v).  \\
- \texttt{scroll [direction=down|up]}: Scroll the page up or down.  \\
- \texttt{new\_tab}: Open a new, empty browser tab.  \\
- \texttt{tab\_focus [tab\_index]}: Switch the browser's focus to a specific tab using its index.  \\
- \texttt{close\_tab}: Close the currently active tab.  \\
- \texttt{goto [url]}: Navigate to a specific URL.  \\
- \texttt{go\_back}: Navigate to the previously viewed page.  \\
- \texttt{go\_forward}: Navigate to the next page (if a previous \texttt{go\_back} action was performed).  \\
- \texttt{stop [answer]}: Issue this action when you believe the task is complete. If the objective is to find a text-based answer, provide the answer in the bracket. If you believe the task is impossible to complete, provide the answer as ``N/A'' in the bracket.\\
\\
**Output Format** \\
First, generate the reasoning process for the action. Then, generate the action in the correct format. Start with a ``In summary, the next action I will perform is'' phrase, followed by action inside \texttt{\`}\texttt{\`}\texttt{\`}. \\
For example: \\
``Let's think step-by-step. To add a product to the shopping cart, I need to navigate to the catalog or product section. 
The \texttt{"CATALOG"} link is available with ID [1234]. 
In summary, the next action I will perform is \texttt{\`}\texttt{\`}\texttt{\`}click [1234]\texttt{\`}\texttt{\`}\texttt{\`}''. \\
\\
\textbf{Instruction:} \{instruction\} \\
\textbf{Accessibility tree:} \{a11y\_tree\} \\
\textbf{Action History:} \{action\_history\} \\
What's the next action?
        }}
    }
    \captionsetup{labelformat=default, name=Prompt}
    \caption{Prompts for evaluating our agents on WebArena.}
    \label{fig:webarena-our-agent-eval-prompt}
\end{figure*}

\begin{figure*}
    \centering
    \setlength{\fboxrule}{0.85pt}
    \fbox{ \footnotesize 
        \parbox{1\textwidth}{\texttt{\textbf{Evaluation Prompt for WebArena}\\
        \\
\texttt{prompt = \{
    ``intro'': \texttt{``}\texttt{``}\texttt{``}You are an autonomous intelligent agent tasked with navigating a web browser. You will be given web-based tasks. These tasks will be accomplished through the use of specific actions you can issue.  
    \\
    Here's the information you'll have:  
    The user's objective: This is the task you're trying to complete.  
    The current web page's accessibility tree: This is a simplified representation of the webpage, providing key information.  
    The current web page's URL: This is the page you're currently navigating.  
    The open tabs: These are the tabs you have open.  
    The previous action: This is the action you just performed. It may be helpful to track your progress.  
    The screenshot of current webpage: This .png image will be input as base64 format and the image is for you to better understand the web page, providing key information.  
\\
    The actions you can perform fall into several categories:  
\\
    \textbf{Page Operation Actions}:  \\
    \texttt{\`}click [id]\texttt{\`}: This action clicks on an element with a specific id on the webpage. Note that you CAN ONLY answer the id (a number) instead of clicking a text like `click [month]'.  \\
    \texttt{\`}type [id] [content] [press\_enter\_after=0|1]\texttt{\`}: Use this to type the content into the field with id. By default, the ``Enter'' key is pressed after typing unless press\_enter\_after is set to 0.  \\
    \texttt{\`}hover [id]\texttt{\`}: Hover over an element with id.  \\
    \texttt{\`}press [key\_comb]\texttt{\`}: Simulates the pressing of a key combination on the keyboard (e.g., Ctrl+v).  \\
    \texttt{\`}scroll [down/up]\texttt{\`}: Scroll the page up or down. You need to output the command like scroll [down] to scroll down. 
\\
    \textbf{Tab Management Actions}:  \\
    \texttt{\`}new\_tab\texttt{\`}: Open a new, empty browser tab.  \\
    \texttt{\`}tab\_focus [tab\_index]\texttt{\`}: Switch the browser's focus to a specific tab using its index.  \\
    \texttt{\`}close\_tab\texttt{\`}: Close the currently active tab.  
\\
    \textbf{URL Navigation Actions}:  \\
    \texttt{\`}goto [url]\texttt{\`}: Navigate to a specific URL.  \\
    \texttt{\`}go\_back\texttt{\`}: Navigate to the previously viewed page.  \\
    \texttt{\`}go\_forward\texttt{\`}: Navigate to the next page (if a previous `go\_back' action was performed).  
\\
    \textbf{Completion Action}:  \\
    \texttt{\`}stop [answer]\texttt{\`}: Issue this action when you believe the task is complete. If the objective is to find a text-based answer, provide the answer in the bracket. If you believe the task is impossible to complete, provide the answer as ``N/A'' in the bracket.  
\\
    \textbf{Homepage}:  
    If you want to visit other websites, check out the homepage at http://homepage.com. It has a list of websites you can visit.  
    http://homepage.com/password.html lists all the account names and passwords for the websites. You can use them to log in to the websites.  
\\
    To be successful, it is very important to follow the following rules:  \\
    1. You should only issue an action that is valid given the current observation.\\  
    2. You should only issue one action at a time.\\  
    3. You should follow the examples to reason step by step and then issue the next action.\\  
    4. Generate the action in the correct format. Start with a ``In summary, the next action I will perform is'' phrase, followed by the action inside \texttt{\`}\texttt{\`}\texttt{\`}\texttt{\`}\texttt{\`}\texttt{\`}. For example,``In summary, the next action I will perform is \texttt{\`}\texttt{\`}\texttt{\`}click [1234]\texttt{\`}\texttt{\`}\texttt{\`}''.\\  
    5. Issue stop action when you think you have achieved the objective. Don't generate anything after stop.'''''' 
\\
    ``examples'': [  
        (  
            ``````OBSERVATION:  
            [1744] link `HP CB782A\#ABA 640 Inkjet Fax Machine (Renewed)',  
            [1749] StaticText '\$279.49',
            [1757] button `Add to Cart', 
            [1760] button `Add to Wish List',  
            [1761] button `Add to Compare',  
            URL: http://onestopmarket.com/office-products/office-electronics.html  
            OBJECTIVE: What is the price of HP Inkjet Fax Machine  
            PREVIOUS ACTION: None'''''',  
            ``Let's think step-by-step. This page lists the information of HP Inkjet Fax Machine, which is the product identified in the objective. Its price is \$279.49. I think I have achieved the objective. I will issue the stop action with the answer. In summary, the next action I will perform is \texttt{\`}\texttt{\`}\texttt{\`}stop [\$279.49]\texttt{\`}\texttt{\`}\texttt{\`}'',  
        ),  
        (  
            ``````OBSERVATION:  
            [164] textbox `Search' focused: True required: False  
            [171] button `Go'  
            [174] link `Find directions between two points'  
            [212] heading `Search Results'  
            [216] button `Close'  
            URL: http://openstreetmap.org  
            OBJECTIVE: Show me the restaurants near CMU  
            PREVIOUS ACTION: None'''''',  
            ``Let's think step-by-step. This page has a search box whose ID is [164]. According to the Nominatim rule of OpenStreetMap, I can search for the restaurants near a location by ``restaurants near''. I can submit my typing by pressing Enter afterwards. In summary, the next action I will perform is \texttt{\`}\texttt{\`}\texttt{\`}type [164] [restaurants near CMU] [1]\texttt{\`}\texttt{\`}\texttt{\`}'',  
        ),  
    ],  
\\
    ``template'': ``````OBSERVATION:  
    {observation},  
    URL: {url},  
    OBJECTIVE: {objective},  
    PREVIOUS ACTION: {previous\_action}'''''',  
\\
    ``meta\_data'': \{  
        ``observation'': ``accessibility\_tree'',  
        ``action\_type'': ``id\_accessibility\_tree'',  
        ``keywords'': [``url'', ``objective'', ``observation'', ``previous\_action''],  
        ``prompt\_constructor'': ``CoTPromptConstructor'',  
        ``answer\_phrase'': ``In summary, the next action I will perform is'',  
        ``action\_splitter'': ``\texttt{\`}\texttt{\`}\texttt{\`}'' 
    \},  
\}}
}}}
    \captionsetup{labelformat=default, name=Prompt}
    \caption{Prompts for evaluating base models (Zero-Shot) on WebArena.}
    \label{fig:webarena-zero-shot-eval-prompt}
\end{figure*}

\begin{figure*}
    \centering
    \setlength{\fboxrule}{0.85pt}
    \fbox{ \footnotesize
        \parbox{1\textwidth}{\texttt{\textbf{Trajectory Reward Model Prompt}\\
        \\
You are an expert in evaluating GUI agent task trajectories. Your task is to assess the quality and effectiveness of task trajectories for GUI manipulation tasks.  
\\
A trajectory consists of the following components:\\
1. High-level Instruction: Describes the user's intended task (e.g., "Create a new blank project name 'OS-Genesis'").\\
2. Action History: Includes two key parts:\\
   - Reasoning and Action for Each Step: A sequence of actions performed by the agent, including the reasoning thought and final executed action.\\
   - GUI Screenshots: Screenshots of the last state: (if there are at least three states; otherwise, include all states).
\\
When evaluating a trajectory, consider these key aspects:
\\
Evaluation Criteria: \\
1. Trajectory Coherence:\\
   - Do the low-level steps and corresponding actions follow a logical sequence toward the goal?\\
   - Are the actions clearly described and specific?\\
   - Are there redundant or unnecessary actions?
\\
2. Task Completion:\\
   - Does the trajectory successfully achieve the instructed task?\\
   - Are all necessary interactions completed?\\
   - Are error cases handled appropriately?
\\
Scoring Guidelines:\\
Rate the trajectory on a scale of 1 to 5 based on the evaluation criteria:
\\
- 5: The task is perfectly completed, successfully executing multiple actions to achieve the goal. The sequence is logically clear with no noticeable redundancies.\\
- 4: The task is mostly completed, successfully executing multiple actions. However, due to challenges or ambiguities in the instructions, the completion is not perfect, or there are inefficiencies in the process.\\
- 3: The task is partially completed, with some successful actions executed. However, due to task or environmental constraints, the goal is not fully achieved, or the sequence ends in a loop or error.\\
- 2: Only a few actions are executed. Although there is an attempt to complete the task, the trajectory deviates from the goal early on or demonstrates significant inefficiencies in execution and logic.\\
- 1: The task fails completely, with no meaningful actions executed at the start. The sequence either falls into an immediate deadlock, a repetitive loop, or demonstrates no value in completing the task. Or the tasks are completely inaccessible.
\\
Note: If the task is relatively complex, but the trajectory demonstrates valuable attempts, even if the task is not fully completed, consider adjusting the score upward. However, if the task is complex but the trajectory fails to perform actions that contribute meaningfully to task completion, no extra points should be awarded.
\\
You need to judge the score based on the agent's actions and screenshots combined.
\\
Response Format:
\\
Format your response into two lines as shown below: 
\\
Reason: <your thoughts and reasoning process for the score>
\\
Score: <your score from 1-5>
        }
    }}
    \captionsetup{labelformat=default, name=Prompt}
    \caption{Prompts for Trajectory Reward Model}
    \label{fig:trm-prompt-mobile}
\end{figure*}
\section{Details about Diversity Analysis}
\label{app:diversity_details}

We visualize the instruction embeddings calculated in Section \ref{sec:diversity_analysis} in Figure~\ref{fig:data_diversity_visualization}. This demonstrates that \ours generates more diverse instructions using an exploration-driven method.

We analyze the average word count in synthesized and human-annotated task instructions. For mobile tasks, Task-Driven and Self-Instruction yield average word counts of 9.64 and 9.84, respectively. In contrast, \ours generates longer instructions with an average of 18.01 words, closely matching the 18.71 words in human data. For web tasks, Task-Driven and Self-Instruction produce averages of 11.79 and 8.45 words, while \ours generates instructions with an average of 19.68 words. These results indicate that \ours produces more detailed instructions with sufficient information and context.

Regarding the average number of steps per task, for mobile tasks, Task-Driven, Self-Instruction, \ours, and Human data have averages of 5.64, 3.43, 5.60, and 5.31 steps, respectively. These are comparable, except that Self-Instruction generates tasks with fewer steps. For web tasks, Task-Driven and Self-Instructions have averages of 8.74 and 7.37 steps, while \ours generates tasks with a shorter average of 4.46 steps.



\end{document}